\documentclass{iopjournal}

\usepackage{hyperref}
\usepackage{natbib}
\usepackage{amsfonts,amsmath,amssymb,amsthm,mathtools, bm}
\usepackage{cleveref}
\usepackage{algorithm}
\usepackage{algpseudocode}
\usepackage{subcaption}

\newcommand{\R}{\mathbb{R}}
\newcommand{\N}{\mathbb{N}}
\DeclareMathOperator{\pe}{\mathbf{PE}}
\DeclareMathOperator{\att}{\mathbf{SA}}
\DeclareMathOperator{\ff}{\mathbf{FF}}
\DeclareMathOperator{\tf}{\mathbf{TF}}
\newcommand{\din}{d_{\text{in}}}
\newcommand{\ip}[2]{\langle #1, #2 \rangle}

\DeclareMathOperator*{\argmin}{arg\,min}

\begin{document}

\articletype{Paper} 

\title{Time-Delayed Transformers for Data-Driven Modeling of Low-Dimensional Dynamics}

\author{Albert Alcalde$^{1,*}$, Markus Widhalm$^2$ and Emre Y{\i}lmaz$^{2}$}

\affil{$^1$Chair for Dynamics, Control, Machine Learning, and Numerics (Alexander von Humboldt Professorship), Friedrich--Alexander--Universit\"at Erlangen--N\"urnberg, 91058 Erlangen, Germany}

\affil{$^2$Institute of Aerodynamics and Flow Technology, DLR, 38108 Braunschweig, Germany}

\affil{$^*$Author to whom any correspondence should be addressed.}

\email{albert.alcalde@fau.de}

\keywords{Transformers, Self-Attention Mechanism, Data-Driven Modeling, Scientific Machine Learning}

\begingroup
\renewcommand{\thefootnote}{}
\footnotetext{Author accepted manuscript. The version of record is
published in \emph{Machine Learning: Science and Technology}
\textbf{7}(4) (2026), 045038.
\href{https://doi.org/10.1088/2632-2153/ae8e2d}{doi:10.1088/2632-2153/ae8e2d}.}
\endgroup

\begin{abstract}
We propose the time-delayed transformer (TD-TF), a simplified transformer architecture for data-driven modeling of unsteady spatio-temporal dynamics. TD-TF bridges linear operator-based methods and deep sequence models by showing that a single-layer, single-head transformer can be interpreted as a nonlinear generalization of time-delayed dynamic mode decomposition. The architecture is deliberately minimal, consisting of one self-attention layer with a single query per prediction and one feedforward layer, resulting in linear computational complexity in sequence length and a small parameter count. Numerical experiments demonstrate that TD-TF matches the performance of linear baselines on near-linear systems, while significantly outperforming them in nonlinear and chaotic regimes, where it accurately captures long-term dynamics. Validation studies on synthetic signals, unsteady aerodynamics, the Lorenz '63 system, and a reaction-diffusion model show that TD-TF preserves the interpretability and efficiency of linear models while providing substantially enhanced expressive power for complex dynamics.
\end{abstract}

\section{Introduction}

The availability of increasingly large high-resolution datasets, both in space and time, together with recent advances in scientific machine learning (SciML), is transforming how we model the evolution of physical systems. In particular, the modeling of unsteady flows and other high-dimensional dynamical systems has benefited from such data-centric approaches, which complement or extend classical methods derived from first principles.

Contrary to the traditional \textit{physics-driven} paradigm, where one derives a system of ordinary differential equations (ODEs) from governing physical laws, the \textit{data-driven} approach \citep{qin2019data,chen2025due} aims to identify an evolution operator that reproduces the observed dynamics directly from data. Even when the measurements are partial or low-dimensional, time-delay embedding techniques \citep{takens1981detecting} allow for reconstructing the underlying attractor and recovering temporal correlations. Koopman operator theory \citep{koopman1931hamiltonian,mezic2005spectral} provides a linear framework for representing nonlinear systems, and serves as the theoretical basis of many modern data-driven models. Within this family, dynamic mode decomposition (DMD) \citep{SCHMID_2010} offers a simple yet effective approach to extract the best-fit linear dynamical system describing the temporal evolution of observables. By enriching the state with delayed snapshots, time-delayed DMD (TD-DMD) captures higher-order temporal structures \citep{tu2013dynamic,leClainche2017higherOrder}. However, TD-DMD remains linear, which fundamentally limits its ability to capture strongly nonlinear or chaotic behavior commonly encountered in physical systems.

At the other end of the modeling spectrum, transformer architectures \citep{vaswani2017attention} have emerged as powerful nonlinear sequence models built from feedforward neural networks \citep{Cybenko1989ApproximationBS,he2016deep} and self-attention mechanisms \citep{vaswani2017attention}. Originally developed for natural language processing, transformers have since been adapted to time-series forecasting \citep{zhou2021informer,wu2021autoformer} and operator learning \citep{cao2021choose,hemmasian2023reduced,calvello2024continuum} in scientific machine learning. The lack of transparency and the quadratic complexity of full attention can be problematic in engineering contexts, where understanding the model and scaling to long sequences are important. By using self-attention to weight past information, transformers can learn complex dependencies across multiple time scales. These models achieve remarkable expressiveness on challenging tasks such as turbulent flow prediction and the construction of surrogate models for partial differential equations (PDEs) \citep{geneva2022transformers,Wu2022ROMTransformer,solera2024beta}. However, general-purpose transformers typically involve deep, multi-layer architectures that sacrifice interpretability and efficiency. 

The tension between interpretability and expressivity is particularly evident in unsteady aerodynamics \citep{bekermeyer2019}. Short-term atmospheric disturbances, known as gust encounters, can cause rapid changes in lift and drag, potentially leading to structural overstressing or loss of control in flight \citep{etkin72}. During aircraft design and certification, large ensembles of such unsteady scenarios must be analyzed to ensure compliance with safety standards \citep{easa07}. High-fidelity CFD simulations are often prohibitively expensive due to the nonlinear phenomena involved (e.g., shocks and flow separation), while simplified linearized models lack accuracy in strongly unsteady regimes \citep{albano69,verdon84}. These challenges highlight the need for efficient yet interpretable data-driven models capable of learning nonlinear temporal responses from limited training data. Motivated by this challenge, we hypothesize that introducing structured nonlinearity into time-delayed models can bridge the gap between linear methods like TD-DMD and black-box sequence models.

Our approach proposes a new structured transformer architecture, the time-delayed 
transformer (TD-TF). TD-TF is deliberately designed to mirror the form of TD-DMD while extending it with nonlinear expressive capacity. In our formulation, the model's prediction at each time step can be interpreted as the action of a learned evolution operator on a finite window of past states. In concrete terms, TD-TF produces the next state as a weighted sum of nonlinear transformations of the past $n$ states: each past snapshot is first passed through a feedforward network which acts as a learned nonlinear \textit{feature map}, and then combined via attention weights that depend on their relevance to the latest state. This preserves the central assumption of TD-DMD that future states depend only on a finite recent history, but relaxes the linearity constraint, allowing those past states to interact in a nonlinear manner.

\subsection{Our contributions}

This work introduces a new perspective on unifying linear delay-based models and transformers. Our main contributions are:
\begin{itemize}
    \item \textbf{Nonlinear reinterpretation of TD-DMD.} We demonstrate that a single-layer, single-head self-attention model can be interpreted as a nonlinear generalization of TD-DMD. This provides a direct theoretical bridge between our transformer architecture and classical data-driven system identification techniques.

    \item \textbf{Novel time-delayed transformer architecture.} We propose a new architecture that enforces a TD-DMD-inspired structure, using one feedforward transformation per state and one attention query per prediction, yielding linear complexity in sequence length. Positional information is encoded by concatenating time indices rather than learned embeddings, further simplifying the model.

    \item \textbf{Numerical validation.} We validate TD-TF on a range of systems from simple linear to highly nonlinear and chaotic. These include (i) sinusoidal signals where an exact linear solution exists, (ii) an unsteady aerodynamics problem (gust response of an airfoil) with high-dimensional flow dynamics close to a linear regime, (iii) the Lorenz '63 chaotic system, and (iv) a reaction--diffusion partial differential equation projected onto a low-dimensional basis. Our experiments show that TD-TF achieves comparable accuracy to TD-DMD in or close to linear regimes, and improves upon TD-DMD in strongly nonlinear and chaotic regimes.
\end{itemize}

\subsection{Related work}

Learning temporal dynamics from data has been studied extensively, from classical linear identification methods \citep{timeSeries2016} to modern neural operator models \citep{lu2021learning}. We briefly review the main approaches most relevant to this work.  

\paragraph{Data-driven dynamical modeling.}  
A large body of work focuses on identifying system dynamics from data. Classical linear techniques such as dynamic mode decomposition (DMD) \citep{SCHMID_2010} and its extensions, such as extended and kernel DMD \citep{williams2015data,williams2014kernel}, approximate the Koopman operator \citep{koopman1931hamiltonian,mezic2005spectral} to obtain linear representations of nonlinear dynamics. Time-delay embeddings \citep{takens1981detecting} enable reconstruction of partially observed systems and lead to methods such as TD-DMD \citep{tu2013dynamic,leClainche2017higherOrder} and HAVOK \citep{brunton2017chaos}. Equation discovery approaches such as sparse identification of nonlinear dynamics (SINDy) \citep{brunton2016discovering,schaeffer2017learningPDEs} recover governing equations through sparse regression, whose interpretability and performance are tied to selecting a proper set of candidate terms for the library.

\paragraph{Transformers for sequence learning.} 
Transformers have been applied to time-series forecasting tasks in many domains. Recent studies draw parallels between self-attention and autoregressive (AR) models \citep{lu2025linear,you2024linear}, where the transformer's attention mechanism with a causal mask is seen as a data-driven generalization of a linear vector autoregressive (VAR) process \citep{lutkepohl2005new}. This connection has been used to interpret transformers as nonlinear AR models and to explore their ability to capture long-range dependencies. More recently, architectures such as the Delayformer have been proposed to model spatiotemporal dynamics by explicitly integrating temporal transformations for multivariate prediction, reflecting growing interest in attention-based models tailored to dynamics \citep{delayformer2026}. 
Relatedly, recent work has adopted an operator-learning viewpoint and shown that transformer models can learn transfer operators in context \citep{bao2026transformers}.
In contrast to these works, our formulation of the TD-TF explicitly enforces an AR structure by attending only from the last input state to previous states, rather than attending over all pairwise interactions.

\paragraph{Transformers for physical systems.}  
Recently, neural operators such as the Fourier Neural Operator \citep{li2021fourier, kovachki2023neural} and DeepONet \citep{lu2021learning} have been proposed for learning solution operators of physical systems governed by PDEs. Recent developments further connect attention, state-space sequence models, and neural-operator ideas for evolving physical systems, including transformer-based forecasting from sparse measurements \citep{song2024forecasting}, attention-based spatio-temporal neural operators \citep{karkaria2025attention}, and state-space neural operators for dynamical systems \citep{hu2025state}. Transformers have been adapted to this setting due to their scalability and capacity to model long-range dependencies, and have demonstrated strong performance relative to convolutional or recurrent architectures on complex PDEs \citep{cao2021choose,calvello2024continuum, herde2024poseidon}. Our work differs by having the objective of retaining interpretability by using a single self-attention layer and a constrained architecture, thus avoiding the black-box nature of deep transformers.

\subsection{Organization of the paper} 

The remainder of the paper is organized as follows. In \Cref{s:probSetting}, we formalize the problem setting. In \Cref{s:modeling}, we introduce and motivate the two delay-based models considered in this work, emphasizing their connections and distinctive features. In \Cref{s:numerics}, we present numerical results on four case studies of increasing complexity, comparing TD-TF with TD-DMD. Finally, we summarize our conclusions in \Cref{s:conclusions}.

\section{Problem setting}\label{s:probSetting}

We consider an autonomous dynamical system described by a system of ordinary differential equations:
\begin{equation}
    \frac{d \mathbf{w}(t)}{dt} = \mathbf{f}(\mathbf{w}(t)), \quad \mathbf{w}(0) = \mathbf{w}_0,
\end{equation}
where the vector-valued function $\mathbf{f}: \R^d \to \R^d$ is unknown. However, we can observe $\mathbf{w} \in \R^d$ using a set of $N$ available trajectories. Along a sequence of equally spaced time instances, $t_k = k \Delta t$, where  $k = 0, \dots, K$, we denote the state at time $t_k$ along the $i$-th trajectory with initial condition $\mathbf{w}_0^i$ as
\begin{equation}
    \mathbf{w}_k^i = \mathbf{w}(t_k; \mathbf{w}_0^i), \quad \text{where} \quad k = 0, \dots, K\quad\text{and}\quad i = 1,\dots, N. \label{eqn_u_evol}
\end{equation}
Our objective is to create a data-driven model that predicts the evolution of $\mathbf{w}$ from any initial state $\mathbf{w}_0^i$. In many real-world scenarios, $\mathbf{w}$ is a low-dimensional observable of a large (and possibly infinite) dimensional system, so we are particularly interested in dynamics where $d$ is relatively small. 

Furthermore, we relax the strong requirement of having access to the observable for all time instances. Following the framework of \citep{chen2025due}, we assume access to only bursts of consecutive time instances randomly spaced along the full trajectories. From this dataset, the dynamics of the observable $\mathbf{w}$ can be reconstructed in the following way.

First, we denote the $J$ bursts of $n+1$ consecutive states in time (randomly sampled from the $N$ total trajectories) by
\begin{equation}
    (\mathbf{w}_0^j, \dots, \mathbf{w}_{n}^j), \quad j = 1, \dots, J.
\end{equation}
Then, a delay-based model, $\mathbf{F}: \R^{nd} \to \R^d$, is built by incorporating information from the previous $n$ states to predict the next state
\begin{equation}
      \mathbf{w}_{n}^j  \approx \mathbf{F}([\mathbf{w}_0^j; \dots; \mathbf{w}_{n-1}^j]), \quad j = 1, \dots, J. 
\end{equation}

\begin{figure}
    \centering
    \includegraphics[width=0.65\linewidth]{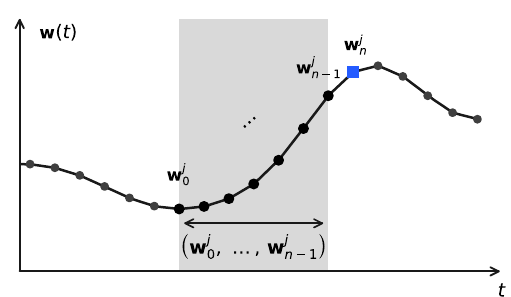}
    \caption{Schematic representation of the time-delay embedding for a one-dimensional trajectory $\mathbf{w}(t)$. A burst of $n$ past consecutive states $(\mathbf{w}_0^j, \dots,\mathbf{w}_{n-1}^j)$ is used to predict the next state $\mathbf{w}_n^j$, marked in blue.}
    \label{fig:1DsignalModified}
\end{figure}

An example of this time-delay embedding of a one-dimensional observable is illustrated in \Cref{fig:1DsignalModified}.
Many physical phenomena fit this problem formulation, such as two-dimensional flow past a cylinder, for which we can measure the lift coefficient, $C_l \in\R$, along certain time windows. In this paper, we consider the scenario of measuring several trajectories of an observable, with the goal of reconstructing trajectories for arbitrary times and initial conditions.

\section{Delay-based data-driven modeling}\label{s:modeling}

In this section, we introduce the two data-driven methods considered in this work, namely, (i) time-delayed dynamic mode decomposition and (ii) the time-delayed transformer. In particular, we highlight how these two methods are related to each other.

\subsection{Time-delayed dynamic mode decomposition}

We start by briefly reviewing dynamic mode decomposition \citep{SCHMID_2010}. The core idea behind this method is to find the best linear approximation such that
\begin{equation}
    \mathbf{w}_{k+1}^j \approx \mathbf{A} \mathbf{w}_k^j
\end{equation}
holds for all times $k = 0, \dots, n-1$ and all $j = 1, \dots, J$, where the linear mapping is represented by the matrix $\mathbf{A}\in \R^{d\times d}$. For a compact representation, we denote $\mathbf{X}$ and $\mathbf{Y}$ as follows:
\begin{align}
\mathbf{X} &= 
\begin{bmatrix}
\mathbf{w}_0^1 & \cdots & \mathbf{w}_{n-1}^1 & \;\big\vert\; & \cdots & \big\vert & \mathbf{w}_0^J & \cdots & \mathbf{w}_{n-1}^J
\end{bmatrix}
\in \mathbb{R}^{d\times nJ}, \\
\mathbf{Y} &= 
\begin{bmatrix}
\mathbf{w}_1^1 & \cdots & \mathbf{w}_{n\phantom{-1}}^1 & \big\vert & \cdots & \;\big\vert\; & \mathbf{w}_1^J & \cdots & \mathbf{w}_{n\phantom{-1}}^J
\end{bmatrix}
\in \mathbb{R}^{d\times nJ}.
\end{align}

In this case, we assume that the approximation $\mathbf{Y} \approx \mathbf{A} \mathbf{X}$ holds, for which the minimum-residual solution in the Frobenius norm is
\begin{equation}\label{eq:dmdArgmin}
    \argmin_{\mathbf{A}} \| \mathbf{Y} - \mathbf{A} \mathbf{X} \|_\mathrm{F}^2 = \argmin_{\mathbf{A}} \sum_{j=1}^{J} \sum_{k=0}^{n - 1} \| \mathbf{w}^j_{k+1} - \mathbf{A} \mathbf{w}^j_k \|^2.
\end{equation}
An approximate solution for the minimization problem corresponding to \eqref{eq:dmdArgmin} can be attained by employing the following expression:
\begin{equation}
    \mathbf{A} = \mathbf{Y} \mathbf{X}^\dagger = \mathbf{Y} \mathbf{V} \mathbf{\Sigma}^\dagger \mathbf{U}^*,
\end{equation}
where $\dagger$ denotes the Moore-Penrose inverse or \textit{pseudoinverse}, and $\mathbf{X} = \mathbf{U} \mathbf{\Sigma} \mathbf{V}^*$ is the singular value decomposition (SVD) of $\mathbf{X}$.
Given an initial state $\mathbf{w}_0 \in \R^d$ and a trajectory index $j$, the dynamics can be reconstructed as 
\begin{equation}
    \mathbf{w}^j_k = \mathbf{A} \mathbf{w}^j_{k-1} = \mathbf{A}^k \mathbf{w}^j_0
\end{equation}
for all $k \geq 0$.

A well-known shortcoming of DMD is that it might fail to reconstruct complex dynamics when the dimension of the observables is relatively low. The scalar case, $d=1$, is particularly restrictive, since the DMD matrix, $\mathbf{A}=a\in \R$, boils down to a scalar. The model then becomes 
\begin{equation}
    \mathbf{w}_{k+1} = a \mathbf{w}_k, \quad k\geq 0,
\end{equation}
with solution $\mathbf{w}_k = a^k \mathbf{w}_0$. Thus, only single-exponential dynamics can be represented, so multi-rate or more complex temporal behavior cannot be captured.

Time-delayed dynamic mode decomposition (TD-DMD) \citep{arbabi2017ergodic} seeks to overcome this problem by enlarging the states to a prescribed number of time-delays, $n\geq 0$, an approach based on Takens' embedding theorem \citep{takens1981detecting}. The idea is that the following sum of linear approximations
\begin{equation}
    \mathbf{w}_{n}^j \approx \mathbf{A}_{n-1} \mathbf{w}_{n-1}^j + \dots + \mathbf{A}_0 \mathbf{w}_0^j
\end{equation}
is assumed to hold approximately for all trajectories $j = 1, \dots, J$. More compactly, we define augmented states for all pairs of indices $i\leq k$ as
\begin{equation}
    \mathbf{w}_{i:k}^j = \begin{bmatrix} \mathbf{w}_i^j \\ \vdots \\ \mathbf{w}_{k}^j \end{bmatrix} \in \R^{d(k - i + 1)}
\end{equation}
and build the matrices of the augmented states as
\begin{align}
    \mathbf{X} &= \begin{bmatrix}
\mathbf{w}_{0:n-1}^1 & \cdots & \mathbf{w}_{0:n-1}^J
\end{bmatrix} \in \R^{dn\times J}  \\
     \mathbf{Y} &= \begin{bmatrix}
\mathbf{w}_{1:n\phantom{-1}}^1 & \cdots & \mathbf{w}_{1:n\phantom{-1}}^J
\end{bmatrix} \in \R^{dn\times J} 
\end{align}
If $\mathbf{Y} \approx \mathbf{A}_\text{TD} \mathbf{X}$ holds, $\mathbf{A}_{\text{TD}}$ has the following structure for all $j = 1, \dots, J$:
\begin{equation}\label{eq:TD-DMD}
    \underbrace{\begin{bmatrix} \mathbf{w}_1^j \\ \mathbf{w}_2^j \\ \vdots \\ \mathbf{w}_{n-1}^j\\ \mathbf{w}_{n}^j \end{bmatrix}}_{\mathbf{w}_{1:n}^j} \approx 
   \underbrace{
   \begin{bmatrix}
      \mathbf{0}_d & \mathbf{I}_d & \mathbf{0}_d   & \cdots & \mathbf{0}_d \\
      \mathbf{0}_d & \mathbf{0}_d & \mathbf{I}_d & \ddots & \vdots \\
      \vdots& \ddots& \ddots& \ddots & \mathbf{0}_d \\
      \mathbf{0}_d & \cdots& \mathbf{0}_d & \mathbf{0}_d & \mathbf{I}_d \\
      \mathbf{A}_0 & \cdots & \cdots & \mathbf{A}_{n-2} & \mathbf{A}_{n-1}
   \end{bmatrix}
   }_{\mathbf{A}_{\text{TD}}}
   \underbrace{\begin{bmatrix} \mathbf{w}_0^j \\ \mathbf{w}_1^j \\ \vdots \\ \mathbf{w}_{n-2}^j\\ \mathbf{w}_{n-1}^j \end{bmatrix}}_{\mathbf{w}_{0:n-1}^j}.
\end{equation}
Here, $\mathbf{I}_d \in \R^{d\times d}$ and $\mathbf{0}_d \in \R^{d\times d}$ denote the identity matrix and the matrix full of zeros, respectively. 
As in DMD, the best-fit linear operator is obtained in the minimal Frobenius norm sense as $\mathbf{A}_{\text{TD}} = \mathbf{Y} \mathbf{X}^\dagger$. The matrix that contains the last $d$ rows of $\mathbf{A}_{\text{TD}}$ can be denoted by 
\begin{equation}
    \mathbf{\hat{A}}_{\text{TD}} = \begin{bmatrix}
    \mathbf{A}_0 & \cdots & \mathbf{A}_{n-1} 
    \end{bmatrix} \in \R^{d \times nd}.
\end{equation}
Then, the future evolution of an initial augmented state, $\mathbf{w}^{j} \in \R^{nd}$, can be generated by recursively applying 
\begin{equation}
    \mathbf{w}_k^j = \mathbf{\hat{A}}_{\text{TD}} \mathbf{w}_{k-n:k-1}^j
\end{equation}
for all $k \geq n$.

The advantages of TD-DMD are that it is computationally cheap, only requiring SVD and $O(n)$ operations at evaluation time, as well as being rooted in Koopman operator theory \citep{tu2013dynamic}. The main limitation stems from the fact that it is a linear approximation. Therefore, it might struggle to capture complex nonlinear dynamics. Furthermore, the number of parameters grows with the number of delays $n$, which is also a limiting factor. We address these limitations of the TD-DMD approach by introducing time-delayed transformers, inspired by the transformer architecture and its similarities to TD-DMD.

\subsection{Time-delayed transformers}

This section introduces the main components of the transformer architecture that we use to build time-delayed transformers, namely, positional encoding, feedforward layers, and self-attention layers.

\paragraph{Positional encoding.}
It is possible that identical state values occur at different time instances, which would prevent the model from distinguishing between them. \textit{Positional encoding} tackles this issue by including explicit temporal information in each state. In our particular implementation, we simply append to each state its normalized time index. This yields
\begin{equation}
    \mathbf{y}_k^j = \pe (\mathbf{w}_k^j) \coloneqq 
    \begin{bmatrix}
        \mathbf{w}_k^j \\
        k / n
    \end{bmatrix}
    \in \R^{\din}, 
    \quad \text{for all} \; \; k = 0, \dots, n-1,
\end{equation}
where $\din = d+1$. This choice is deliberately minimal: self-attention without positional information is permutation equivariant, and appending the normalized time index breaks this symmetry while avoiding the additional architectural complexity of sinusoidal or learned positional embeddings \cite{vaswani2017attention,Yun2020Are}.

\paragraph{Feedforward layer.}
The first key element of a transformer is the feedforward layer. This is defined as a map $\ff: \R^{\din} \to \R^{\din}$ shared across time steps, and is parametrized as a shallow fully-connected neural network. For a given state, $\mathbf{y}\in \R^{\din}$, this mapping can be expressed as
\begin{equation}
    \ff(\mathbf{y}) = \mathbf{W} \boldsymbol\sigma (\mathbf{U} \mathbf{y} + \mathbf{b}),
\end{equation}
where $\boldsymbol\sigma: \R^{h} \to \R^{h}$ is a nonlinear activation function, $\mathbf{W} \in \R^{\din\times h}$, $\mathbf{U} \in \R^{h\times \din}$ are the matrices of the two consecutive linear layers, and $\mathbf{b} \in \R^h$ is the bias of the first linear layer for the hidden dimension $h\in \N$. 

\paragraph{Self-attention layer.}
The second key element of a transformer is a self-attention layer, defined as the map $\att: \R^{\din n} \rightarrow \R^{d}$ from a collection of $n$ vectors in $\R^{\din}$ to a vector in $\R^{d}$. Specifically, for a given input, $\mathbf{y}_{0:n-1} = [\mathbf{y}_0; \dots; \mathbf{y}_{n-1}] \in \R^{\din n}$, this mapping can be represented as
\begin{align}\label{eq:SA}
\att(\mathbf{y}_{0:n-1}) 
= \sum_{k = 0}^{n-1} \alpha_{n-1,k}(\mathbf{y}_{0:n-1}) \mathbf{V}\mathbf{y}_k 
\end{align}
where
\begin{equation}\label{eq:attentionCoefficients}
    \alpha_{n-1,k}(\mathbf{y}_{0:n-1}) = 
    \frac{ e^{\ip{ \mathbf{y}_{n-1}}{ \mathbf{B}\mathbf{y}_k} }}{ \sum\limits_{r = 0}^{n-1} e^{ \ip{ \mathbf{y}_{n-1}}{\mathbf{B}\mathbf{y}_r} } } \in (0,1).
\end{equation}
Here, $\mathbf{B} \in \R^{\din \times \din}$ and $\mathbf{V}\in \R^{d \times \din}$ are parameter matrices, known as the \textit{query-key} matrix and the \textit{value} matrix, respectively. It is worth noting that the function $\att$ can be applied directly to the augmented states, $\mathbf{y}_{0:n-1}^j \in \R^{\din n}$, or to their transformation after applying the feedforward layer component-wise. We also note that the attention coefficients, $\alpha_{n-1,k}(\mathbf{y}_{0:n-1})$, in \eqref{eq:attentionCoefficients} depend on the pairwise inner-products between the last state, $\mathbf{y}_{n-1}$, and all states in $\mathbf{y}_{0:n-1}$.

\paragraph{Building the time-delayed transformer.}

We define the transformer model by composing positional encoding, a feedforward layer, and a self-attention layer. Specifically, the transformer model is defined as the mapping $\tf: \R^{\din n} \to \R^d$. For a given input already with positional encoding $[\mathbf{y}_0; \dots; \mathbf{y}_{n-1}] \in \R^{\din n}$, this mapping is represented as
\begin{equation}\label{eq:timeDelayedTF}
    \tf\big([\mathbf{y}_0; \dots; \mathbf{y}_{n-1}]\big) = 
    \att \Big(\big[\ff\big(\mathbf{y}_0\big); \dots; \ff\big(\mathbf{y}_{n-1}\big)\big]\Big). 
\end{equation}
We provide a pseudocode description of the TD-TF approach in \Cref{algo:TDTF}. The main benefit of the time-delayed transformer is that it is nonlinear and independent of the number of delays, $n$, in the parameter count, since the same feature maps are used for all delays. However, this comes at the cost of $\mathcal{O}(n)$ operations in the self-attention layer. Therefore, the nonlinearities arising from both the feature maps and the attention weights overcome the limitations of an inherently linear model such as TD-DMD at the cost of expensive training of the parameters in its layers. 

\begin{algorithm}
\caption{Time-delayed transformer}
\begin{algorithmic}[1]\label{algo:TDTF}
\Require States $\{\mathbf{w}_k\}_{k=0}^{n-1} \subset \R^d$; flag \texttt{pos\_enc} $\in \{\texttt{True}, \texttt{False}\}$
\Ensure Output $\mathbf{o}\in\R^{d}$
\Statex
\Function{TD-TF}{$\{\mathbf{w}_k\}_{k=0}^{n-1}$, \texttt{pos\_enc}}
  \If{\texttt{pos\_enc}}
      \State $\din \gets d+1$
  \Else
      \State $\din \gets d$
  \EndIf
  \Statex
  \textbf{Parameters:} $\mathbf{U}\in\R^{h\times\din}$, $\mathbf{W}\in\R^{\din\times h}$, $\mathbf{b}\in\R^{h}$, $\mathbf{B}\in\R^{\din\times\din}$, $\mathbf{V}\in\R^{d\times\din}$; nonlinearity $\boldsymbol{\sigma}:\R^{h}\to\R^{h}$
  \Statex
  \For{$k=0,\dots,n-1$}
     \If{\texttt{pos\_enc}}
         \State $\mathbf{y}_k \gets \begin{bmatrix}\mathbf{w}_k \\ k/n\end{bmatrix} \in \R^{d+1}$
     \Else
         \State $\mathbf{y}_k \gets \mathbf{w}_k \in \R^{d}$
     \EndIf
     \State $\mathbf{z}_k \gets \mathbf{W}\,\boldsymbol{\sigma}(\mathbf{U}\,\mathbf{y}_k + \mathbf{b})$
  \EndFor
  \Statex
  \For{$k=0,\dots,n-1$}
     \State $s_k \gets \langle \mathbf{z}_{n-1},\,\mathbf{B}\,\mathbf{z}_k \rangle$
  \EndFor
  \State $\alpha_{n-1,k} \gets \exp(s_k)\Big/\sum_{r=0}^{n-1}\exp(s_r)$ \quad for $k=0,\dots,n-1$
  \State \Return $\mathbf{o} \gets \sum_{k=0}^{n-1}\alpha_{n-1,k}\,\mathbf{V}\,\mathbf{z}_k$ 
\EndFunction
\end{algorithmic}
\end{algorithm}

\paragraph{Training strategy and evaluation.}

We now describe how the time-delayed transformer is applied to the problem setting described in \Cref{s:probSetting}. Inspired by residual formulations of neural networks \citep{he2016deep}, we include an identity mapping so that the time-delayed transformer learns state increments.
For this, we approximate 
\begin{equation}
    \mathbf{w}_n^j - \mathbf{w}_{n-1}^j \approx \tf([\mathbf{w}_0^j; \dots; \mathbf{w}_{n-1}^j])
\end{equation}
for all $j = 1, \dots, J$. Unlike the TD-DMD approach, for which a closed form solution exists, we need to solve an optimization problem to obtain the model parameters in this case. Next, we denote all parameters in TD-TF as $\boldsymbol\theta = \{ \mathbf{W}, \mathbf{U}, \mathbf{b}, \mathbf{B}, \mathbf{V}\}$, which can be learned by solving the following minimization problem:
\begin{equation}\label{loss:TD-TF}
    \arg\min_{\bm{\theta}} \frac{1}{J} \sum_{j=1}^J \left\| \mathbf{w}_{n}^j - \mathbf{w}_{n-1}^j - \tf_{\boldsymbol{\theta}}([\mathbf{w}_0^j;\dots; \mathbf{w}_{n-1}^j]) \right\|^2.
\end{equation}
As is common with neural network training, we use a stochastic gradient-based algorithm, AdamW \citep{loshchilov2018decoupled}. After training the TD-TF model, it can iteratively be used to reconstruct the dynamics of any initial augmented state $\mathbf{w}_{0:n-1}^{j} \in \R^{nd}$ by recursively applying
\begin{equation}
    \mathbf{w}_k^j = \mathbf{w}_{k-1}^j + 
    \tf([\mathbf{w}_{k-n}^j; \dots; \mathbf{w}_{k-1}^j]) \quad \text{for all  } k \geq n.
\end{equation}

\subsection{Connections to transformer modeling and TD-DMD} \label{ss:relationTDTF}

TD-TF is a simplified variant of the standard transformer architecture, designed to better align with the structure and interpretability of linear delay-based dynamical models. While conventional transformers were developed for high-dimensional sequence tasks such as language and vision, the TD-TF approach isolates the minimal architectural elements required for temporal prediction \citep{zhou2021informer, wu2021autoformer} and reformulates them in a delay-based manner \citep{SCHMID_2010}.

\begin{figure}
    \centering
    \includegraphics[width=0.8\linewidth]{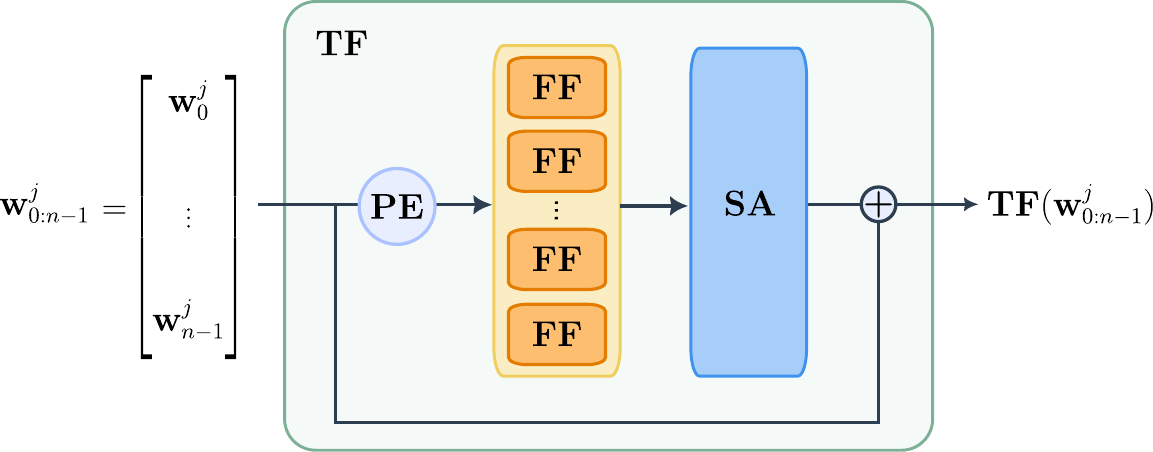}
    \caption{Block diagram of the time-delayed transformer.}
    \label{fig:sketchTDTF}
\end{figure}

As illustrated in \Cref{fig:sketchTDTF}, the TD-TF model consists of a single feedforward layer applied to time-augmented states via positional encoding, followed by one self-attention layer. In contrast to the multi-layer, multi-head structure of standard transformers, we restrict the model to a single attention head and a single transformer block. Motivated by the connection to TD-DMD, we modify the layer ordering by placing the feedforward layer before attention. We also omit normalization layers and replace intermediate residual connections with a single residual link at the model level. Positional information is introduced not through sinusoidal \citep{vaswani2017attention} or learned embeddings \citep{su2024roformer}, but by concatenating a normalized time coordinate to each state vector. 

A key distinction of TD-TF lies in how attention is used for prediction. In standard transformers, self-attention maps $\R^{nd}$ to $\R^{nd}$, computing all pairwise interactions between states at $\mathcal{O}(n^2)$ cost. In contrast, TD-TF evaluates attention only at the final index to predict the next observable, $\mathbf{w}_n$, conditioned on the history $(\mathbf{w}_0, \dots, \mathbf{w}_{n-1})$. This reduces complexity to $\mathcal{O}(n)$, which can mitigate error accumulation during autoregressive rollout. Conceptually, this design brings TD-TF closer to the structure of time-delayed dynamical systems, where each new state is determined by a weighted combination of the past states. Moreover, since our problem setting involves fixed-length input sequences, the model need not generalize across variable-length contexts as in language modeling.

Finally, we establish the direct correspondence that emerges between the TD-TF and TD-DMD models. The TD-DMD update,
\begin{equation}
    \mathbf{w}_n = \sum_{k=0}^{n-1}\mathbf{A}_k\,\mathbf{w}_k,
\end{equation}
can be viewed as the linear analogue of the TD-TF mapping, $\tf(\mathbf{w}_{0:n-1})$, where each term $\mathbf{A}_k\mathbf{w}_k$ is replaced by a nonlinear, data-adaptive contribution,
\begin{equation}
    \alpha_{n-1,k}(\mathbf{w}_{0:n-1})\,\mathbf{V}\,\ff(\mathbf{w}_k).
\end{equation}
In this sense, TD-TF serves as a nonlinear, data-dependent generalization of TD-DMD, where the static matrices $\mathbf{A}_k$ are replaced by adaptive attention weights, and the feature map $\ff$ introduces nonlinear observables.

\section{Numerical experiments}\label{s:numerics}

We evaluate the performance of the proposed TD-TF model against TD-DMD and an additional nonlinear neural baseline across several datasets of increasing dynamical complexity, namely, a simple periodic signal, an unsteady aerodynamic flow, a chaotic nonlinear system and a reaction-diffusion system. All datasets undergo identical preprocessing, where each time series is subsampled every $\tau \in \N$ steps and normalized to the hypercube $[-1,1]^d$. Normalization is inverted at inference time. 

The additional nonlinear baseline is a time-delayed MLP (TD-MLP), which is introduced to distinguish the effect of the TD-TF attention mechanism from the effect of adding nonlinear sequence-modeling capacity. This model is designed to preserve the same delay window as TD-TF but without the attention mechanism. Given the delayed states $\{\mathbf{w}_k\}_{k=0}^{n-1}$, we first optionally apply the same positional encoding as in TD-TF and denote the corresponding input dimension by $\din$. The TD-MLP is then defined as a shared nonlinear map $\Tilde{\ff}:\mathbb{R}^{\din}\to\mathbb{R}^{d}$ applied independently to each delayed state, followed by summation over the delay window:
\begin{equation}
    \mathbf{MLP}
    \big([\mathbf{y}_0;\dots;\mathbf{y}_{n-1}]\big)
    =
    \sum_{k=0}^{n-1}
    \Tilde{\ff}(\mathbf{y}_k),
\end{equation}

The code used to generate the results can be found in the following repository: 
\url{https://github.com/albertalcalde/TDTF}.

\subsection{Case 1: Sinusoidal trajectory}

We consider a generic example of a single sinusoidal trajectory \citep{data-drivenJB2024}. This analytically tractable case provides a baseline for assessing whether the models recover simple harmonic dynamics.

\paragraph{Setup.}

The observable states are defined by the discrete and scalar ($d = 1$) signal, 
\begin{equation}
    w_k = \sin (k\Delta t), \quad k\in\{0,\dots,K\}, \quad \Delta t = \frac{4\pi}{100},
\end{equation}
with $K=200$, $J = 10$ bursts and $\tau = 1$ (i.e., no subsampling). This simple case serves as a benchmark where TD-DMD captures the dynamics for $n=2$ exactly. Indeed, using trigonometric identities, one can show that perfect reconstruction is achieved with $A_1 = 2 \cos(\Delta t)$ and $A_0 = -1$.

\paragraph{Results.}

\begin{figure}
    \centering
    \includegraphics[width= 0.7\linewidth]{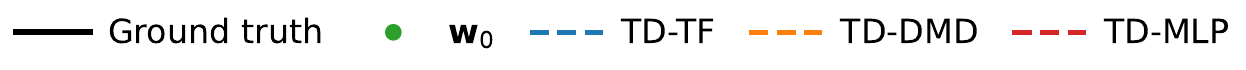} \\
    \includegraphics[width=0.6\linewidth]{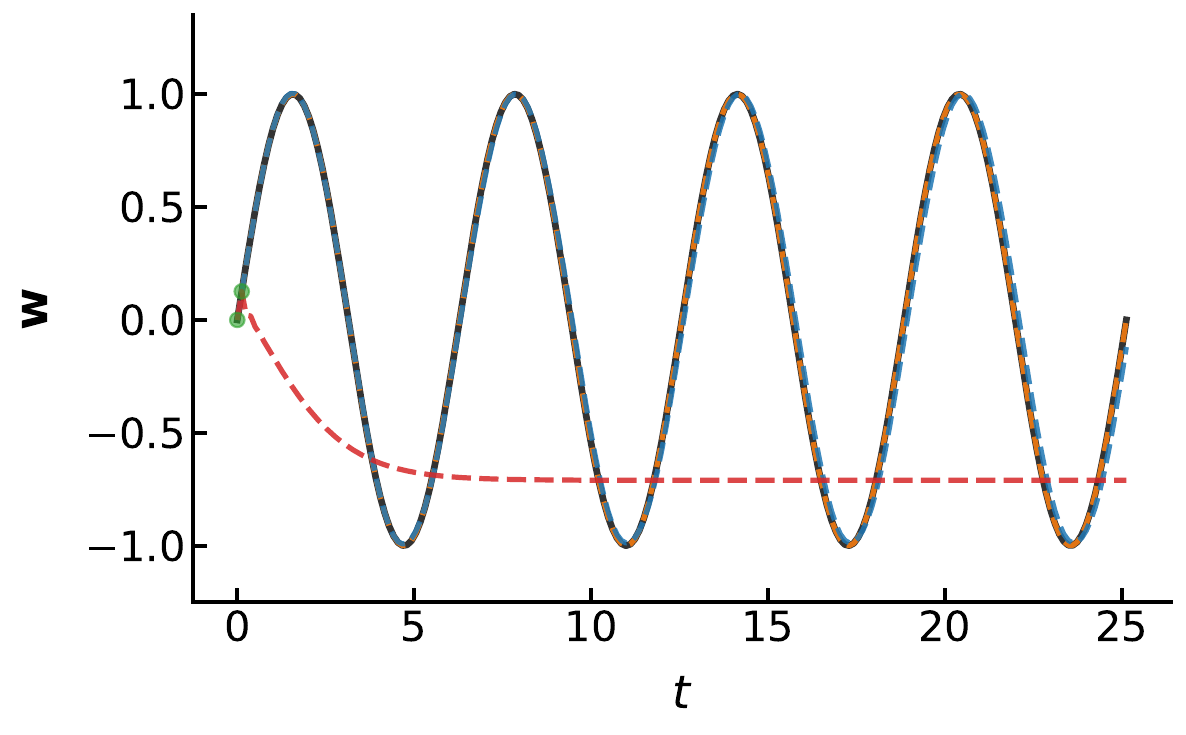}
    \caption{Predictions of the TD-TF, TD-DMD and TD-MLP models for a single sinusoidal trajectory with $n = 2$.}
    \label{fig:sinus}
\end{figure}

For TD-TF, we select the hidden dimension as $10$, the learning rate as $10^{-2}$, and the batch size as $5$, and we train for $1000$ epochs. As shown in \Cref{fig:sinus}, TD-TF can reconstruct the dynamics with $n=2$ delays, but with a noticeable phase shift as time increases further. This qualitative observation is reflected quantitatively by the root mean square error (RMSE): TD-DMD achieves a RMSE of $9.3\cdot 10^{-14}$, corresponding to near machine-precision reconstruction, while TD-TF attains a RMSE of $4.8\cdot 10^{-2}$.

We further compare TD-TF with the TD-MLP baseline using the same hidden dimension and number of delays, and evaluate their performance over $5$ random seeds in \Cref{tab:sinus}. As expected, TD-DMD reconstructs the sinusoidal trajectory to near machine precision, since the signal satisfies an exact second-order linear recurrence. In this linear setting, the nonlinear models do not improve upon TD-DMD, and TD-MLP has the largest error.

\begin{table}
\centering
\caption{RMSE for the sinusoidal trajectory with delay length $n=2$. Neural network models use hidden dimension $h=10$ and are evaluated over $5$ random seeds.}
\begin{tabular}{lccc}
\hline
& TD-TF & TD-DMD & TD-MLP \\ \hline
RMSE & $2.72\cdot 10^{-1} \pm 3.92\cdot 10^{-1}$ & $9.31\cdot 10^{-14}$ & $9.36\cdot 10^{-1} \pm 2.36\cdot 10^{-3}$ \\
\hline
\end{tabular}
\label{tab:sinus}
\end{table}

\subsection{Case 2: Flow around the NLR7301 airfoil}

We next consider unsteady flow around the NLR7301 airfoil \citep{zwaan82} during a vertical gust encounter. This benchmark represents a high-dimensional, nonlinear, and time-dependent aerodynamic regime suitable for evaluating data-driven temporal models.

\paragraph{Setup.}

We generate the flow solutions and quantities of interest using the TAU code, the DLR in-house CFD solver \citep{schwamborn06}, solving the unsteady Reynolds-averaged Navier-Stokes (RANS) equations with a suitable turbulence closure. The unsteady forcing is introduced through a vertical ``$1-\cos$'' gust profile following the discrete gust design criteria \citep{easa07}. The discrete gust design criteria define scenarios by varying parameters including the angle of attack (AoA), free-stream Mach number ($M_{\infty}$), Reynolds number (Re), gust amplitude ($A_g$), and gust length ($l_g$) within prescribed ranges.

For the present demonstration, we fix $M_{\infty}$ at $0.35$ and Re at $1.19 \cdot 10^6$, corresponding to a subsonic flow condition. The gust profile is characterized by a wavelength of $58.0$ m, an amplitude of $8.0$ m/s, and a horizontal offset of $0.2$ m ahead of the airfoil nose. 
The angle of attack is varied in $\{-5, -4, \dots, 10\}$ degrees, yielding $N = 16$ trajectories of aerodynamic coefficients over the physical time domain of $[0, 0.966]$ seconds. Each trajectory is resolved with $K = 1349$ time steps and a temporal resolution of $\Delta t = 7.16 \cdot 10^{-4}$ seconds, and we set $J = 5000$ and $\tau = 30$.

\begin{figure}[h]
    \centering
    \begin{subfigure}{0.49\textwidth}
        \includegraphics[width=\linewidth]{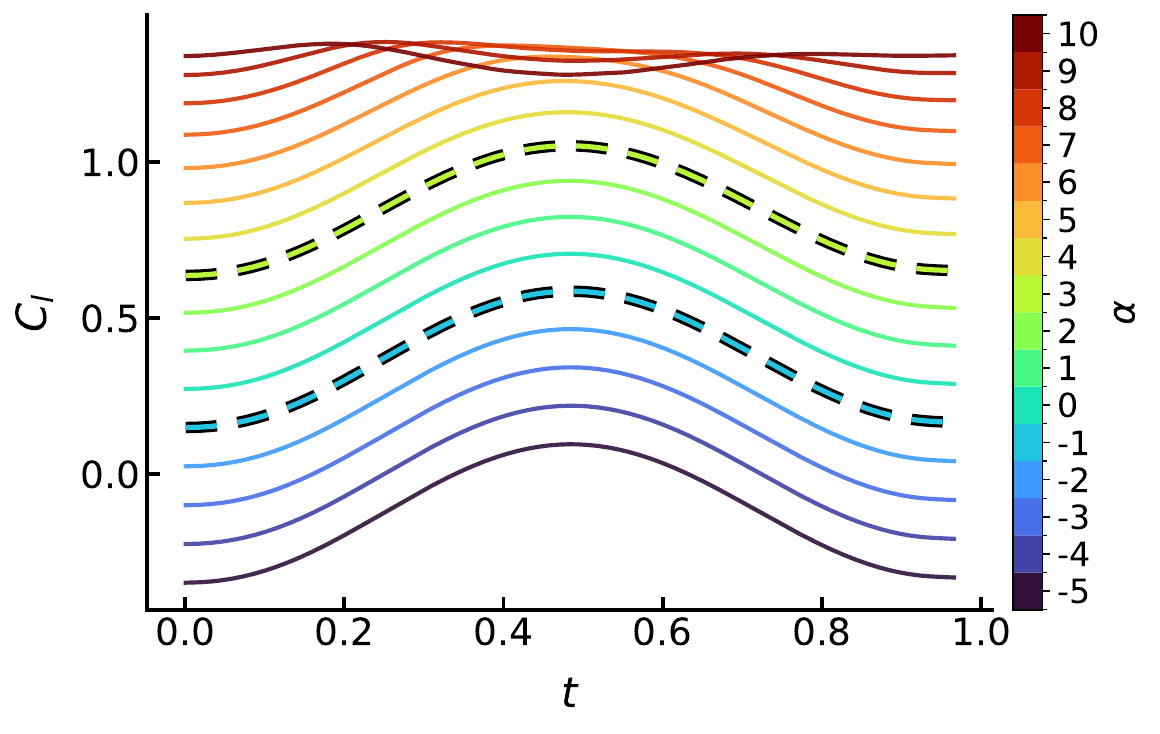}
        \caption{Lift coefficient}
        \label{fig:nlr_lift}
    \end{subfigure}\hfill
    \begin{subfigure}{0.49\textwidth}
        \includegraphics[width=\linewidth]{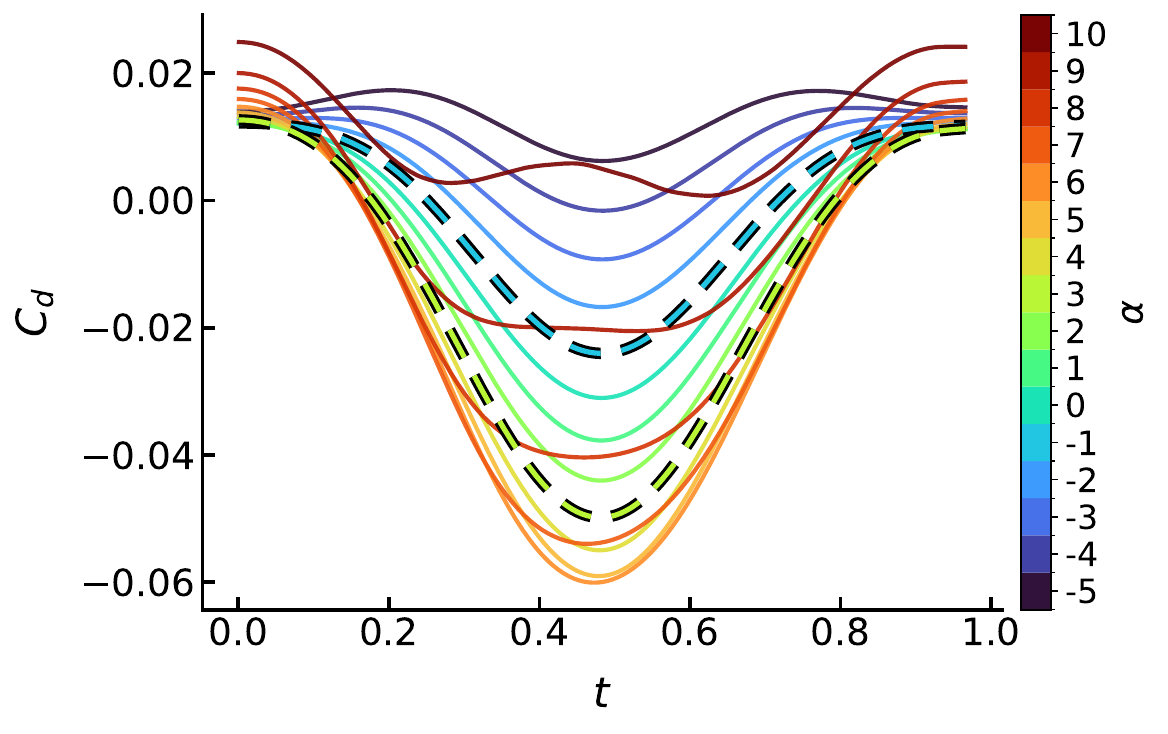}
        \caption{Drag coefficient}
        \label{fig:nlr_drag}
    \end{subfigure}
    \caption{Trajectories in time of lift and drag coefficients of the NLR7301 airfoil case for $\alpha \in \{-5, -4, \dots, 10\}$. The two trajectories with $\alpha = -1^\circ$ and $\alpha = 3^\circ$ are highlighted with dashed lines and black borders.}
    \label{fig:nlr_liftAndDrag}
\end{figure}

The quantities of interest are the lift and drag coefficients, denoted $C_l$ and $C_d$, respectively. In all experiments below, we treat $C_l$ and $C_d$ as \emph{separate scalar observables}: we train one model on the $C_l$ time series and a second, independent model on the $C_d$ time series (i.e., $d=1$ in each case). These time series serve as ground truth and are shown in \Cref{fig:nlr_liftAndDrag}. 

\begin{figure}[h]
    \centering
    \includegraphics[width = 0.7\textwidth]{figs/sinus_legend.pdf}
    \begin{subfigure}{0.49\linewidth}
        \centering
        \includegraphics[width=\linewidth]{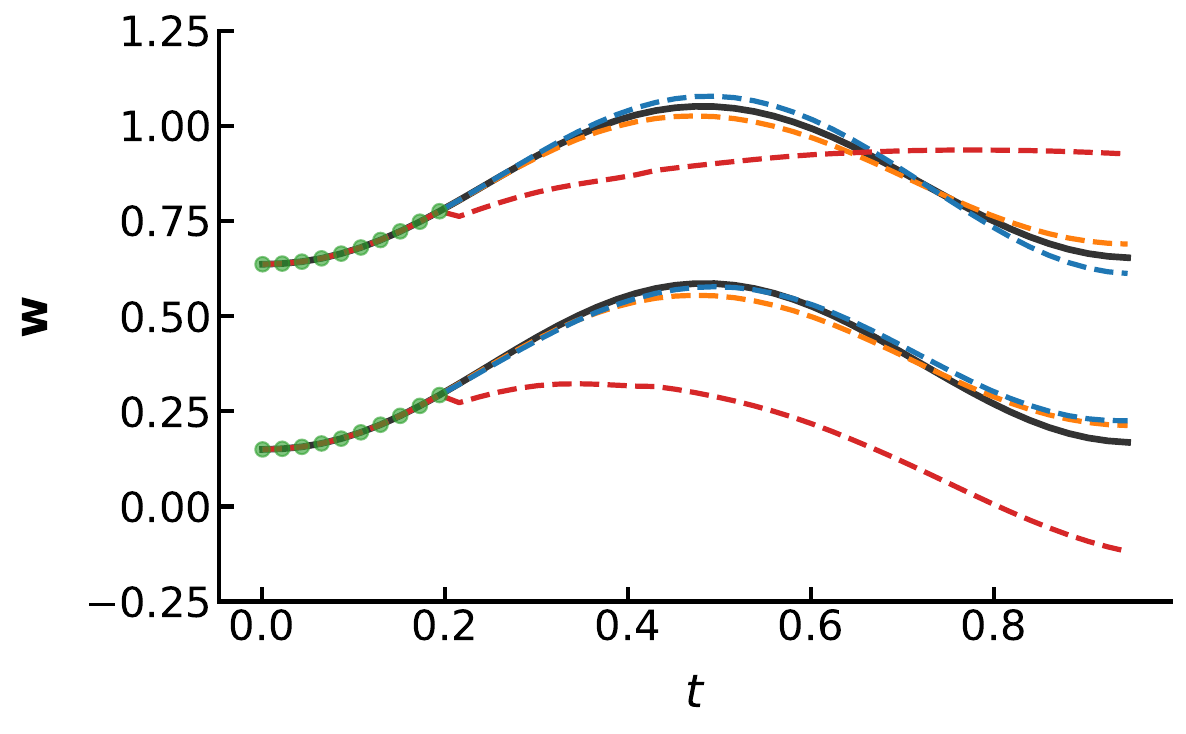}
        \subcaption{Lift coefficient}
        \label{fig:lift_n=10}
    \end{subfigure}\hfill
    \begin{subfigure}{0.49\linewidth}
        \centering
        \includegraphics[width=\linewidth]{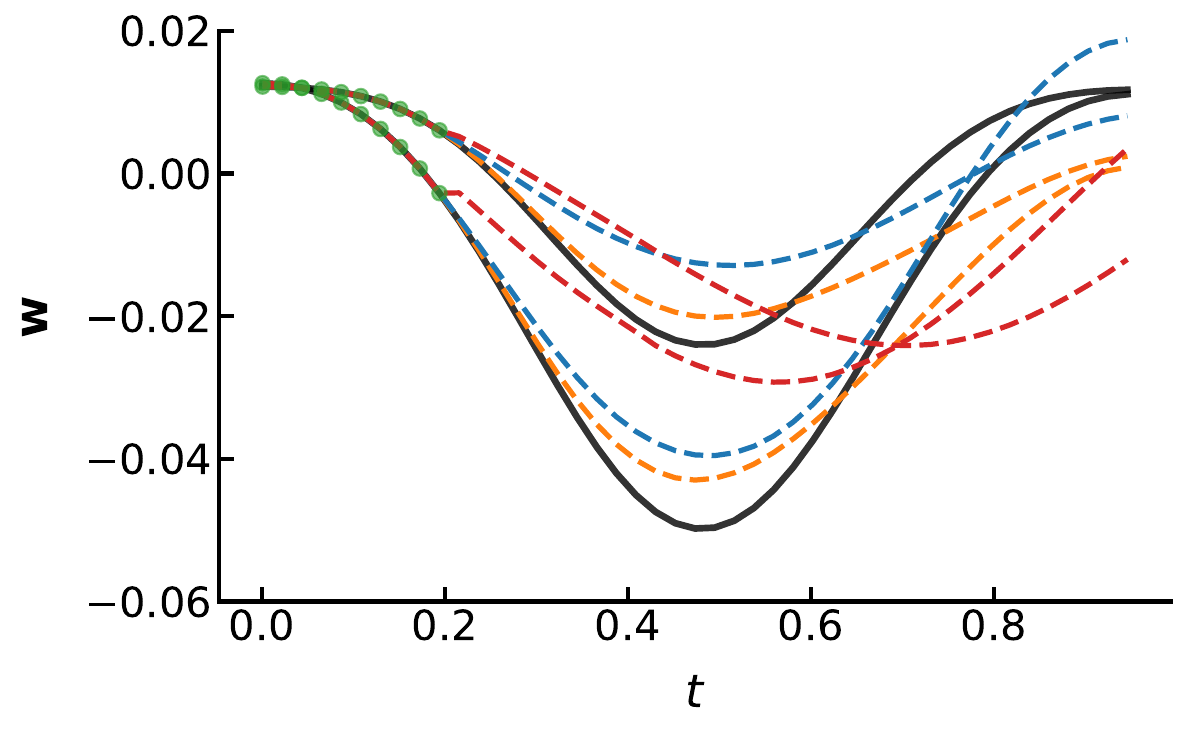}
        \subcaption{Drag coefficient}
        \label{fig:drag_n=10}
    \end{subfigure}
    \caption{Predictions of the TD-DMD, TD-TF and TD-MLP models for the test trajectories $\alpha \in \{-1, 3\}$.}
    \label{fig:nlrPredictions_n=10}
\end{figure}

\paragraph{Results.} 
As a first interpolation test, all trajectories except those corresponding to $\alpha = -1^\circ$ and $\alpha = 3^\circ$ are used as the training set; these two are selected because they fall well within the range of angles considered and are reserved for testing. We train the TD-TF and the TD-MLP models for $500$ epochs with a hidden dimension of $100$, learning rate of $10^{-2}$, and batch size of $100$. We present our results for $n=10$ in \Cref{fig:nlrPredictions_n=10}. 

For the lift coefficient, both the TD-DMD and TD-TF models achieve low average prediction error. In particular, TD-TF attains an average RMSE of $2.03\cdot 10^{-2}$, with a maximum absolute error of $4.9\cdot 10^{-2}$ over the prediction horizon. The drag coefficient presents a more challenging prediction task. TD-TF achieves an average RMSE of $5.57\cdot 10^{-3}$ and a worst-case absolute error of $1.14\cdot 10^{-2}$ over the test trajectories, with errors again increasing toward the end of the prediction horizon. Despite accurately capturing the overall dynamics, TD-TF systematically overestimates the minimum drag value: the predicted minimum is higher than the true minimum by $1.06\cdot 10^{-2}$ on average. Moreover, the time at which the minimum drag occurs is shifted by approximately $3.2\cdot 10^{-2}$ in nondimensional time, consistent with the visual discrepancy observed around $t \approx 0.5$ in \Cref{fig:nlrPredictions_n=10}. 

To assess robustness with respect to initialization, we report RMSE values averaged over $5$ random seeds in \Cref{tab:nlr_lift_drag}. The averaged results show that TD-MLP is less competitive in this example, yielding larger errors than both TD-DMD and TD-TF for both lift and drag. They also show that TD-TF is more sensitive to initialization than TD-DMD in this near-linear aerodynamic setting. These results indicate that TD-TF can be competitive with TD-DMD for favorable initializations, but that the linear delay model remains more robust in this near-linear aerodynamic example. The drag dynamics remain challenging for all models because accurately reproducing sharp extrema requires both amplitude and timing accuracy.

\begin{table}[h]
\centering
\caption{RMSE for the lift and drag trajectories with delay length $n=10$. Both TD-TF and TD-MLP use hidden dimension $h=100$ and are evaluated over $5$ random seeds.}
\begin{tabular}{lccc}
\hline
& TD-TF & TD-DMD & TD-MLP \\ \hline
RMSE (lift) 
& $9.40\cdot 10^{-2} \pm 8.84\cdot 10^{-2}$ 
& $1.95\cdot 10^{-2} \pm 2.71\cdot 10^{-3}$ 
& $1.80\cdot 10^{-1} \pm 4.90\cdot 10^{-2}$ \\
RMSE (drag) 
& $7.85\cdot 10^{-3} \pm 3.77\cdot 10^{-3}$ 
& $6.11\cdot 10^{-3} \pm 1.17\cdot 10^{-4}$ 
& $1.44\cdot 10^{-2} \pm 1.67\cdot 10^{-3}$ \\
\hline
\end{tabular}
\label{tab:nlr_lift_drag}
\end{table}

\begin{figure}[h]
    \centering
    \begin{subfigure}{0.48\linewidth}
        \centering
        \includegraphics[width=\linewidth]{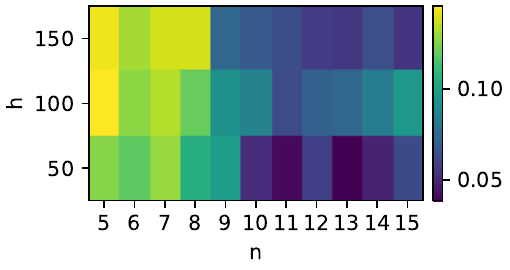}
        \caption{Lift coefficient}
        \label{fig:nlr_extrap_lift_rmse_TDDMD}
    \end{subfigure}\hfill
    \begin{subfigure}{0.48\linewidth}
        \centering
        \includegraphics[width=\linewidth]{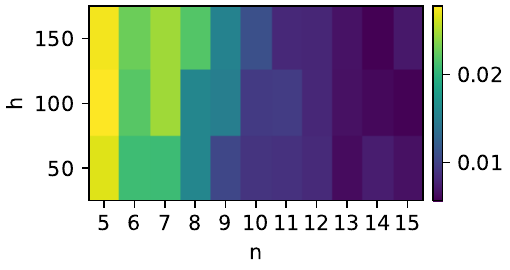}
        \caption{Drag coefficient}
        \label{fig:nlr_extrap_drag_rmse_TDTF}
    \end{subfigure}
    \caption{
    RMSE of the lift and drag coefficient predictions achieved by the TD-TF model in the extrapolation setting (training on $\alpha\in\{-5,\dots,8\}$, testing on $\alpha\in\{9,10\}$), shown as a function of the delay length $n$ and hidden dimension $h$.}
    \label{fig:nlr_extrap_drag_rmse_heatmaps}
\end{figure}

Next, we evaluate the extrapolation capabilities of TD-TF in angle of attack by training on trajectories with $\alpha\in\{-5^\circ ,\dots,8^\circ \}$ and testing on the unseen cases $\alpha=9^\circ$ and $\alpha=10^\circ$. For TD-TF, the dependence of the RMSE on the delay length $n$ and hidden dimension $h$ is shown in \Cref{fig:nlr_extrap_drag_rmse_heatmaps}. We note that, for TD-TF, we account for initialization variance by averaging the results across $5$ training rounds.

For the lift coefficient, TD-TF exhibits large extrapolation errors for small delays ($n\leq 8$), with RMSE values exceeding $1.1\cdot 10^{-1}$ across all tested hidden dimensions and reaching up to approximately $1.4\cdot 10^{-1}$. As the delay length increases, the error decreases substantially, with the minimum RMSE of about $3.9\cdot 10^{-2}$ attained at $n=13$. Comparable performance is observed for $n=11$, where the RMSE drops to roughly $4.1\cdot 10^{-2}$. Increasing the delay length beyond this range does not yield further improvement and can even slightly degrade performance. The RMSE varies noticeably with the hidden dimension $h$, with the smaller value $h = 50$ tending to yield best performance across delay length.

For the drag coefficient, the extrapolation error decreases rapidly with increasing delay length. For $n\geq 10$, the RMSE falls below $9\cdot 10^{-3}$ and continues to decrease with $n$, reaching values close to $5.7\cdot 10^{-3}$ at $n=14$ and remaining at the same order for $n=15$. In this regime, the RMSE shows only weak sensitivity to the hidden dimension $h$, with variations on the order of $10^{-3}$ or less across the tested configurations.

For reference, TD-DMD yields RMSE values in the range $[2.5\cdot 10^{-2},9.4\cdot 10^{-2}]$ for the lift coefficient and $[5.6\cdot 10^{-3},2.3\cdot 10^{-2}]$ for the drag coefficient over the same set of delay lengths, with the minimum errors attained at the largest tested delay, $n=15$. This yields, as in the interpolation case, comparative performance with TD-TF. Notably, in the lift prediction extrapolation case, the linear TD-DMD model attained a lower minimum error than TD-TF, which needed a larger $n$ to reduce error. This suggests that when the true behavior with respect to the parameter is close to the linear regime, a linear model can extrapolate more reliably. TD-TF, being more expressive, might need careful regularization or more data to avoid overfitting for extrapolation.

\subsection{Case 3: Lorenz '63 system}

As the next demonstration case, we examine a canonical chaotic system to assess long-term predictive behavior under strongly nonlinear dynamics.

\paragraph{Setup.}

The Lorenz '63 system \citep{lorenz1963} is a classical chaotic model for atmospheric convection, given by 
\begin{equation}\label{eq:lorenz63}
\begin{cases}
\frac{dx}{dt} &= \sigma (y - x), \\
\frac{dy}{dt} &= x(\rho - z) - y, \\
\frac{dz}{dt} &= xy - \beta z,
\end{cases}
\end{equation}
with $\sigma = 10$, $\rho = 28$, and $\beta = 8/3$.  
We generate $N = 1000$ trajectories using $\Delta t = 10^{-2}$ over $[0, 100]$. To remove transients, we discard the first half of each trajectory, which leaves us with trajectories with $K = 5000$ timesteps. We select $900$ trajectories for training, and leave the remaining $100$ for testing. We retain only the $x$-component of each trajectory, select $J = 5000$ bursts and subsample with $\tau = 16$.

\paragraph{Results.}

\begin{figure}
    \centering
    \includegraphics[width = 0.7\textwidth]{figs/sinus_legend.pdf}
    \begin{subfigure}{0.49\textwidth}
        \centering
        \includegraphics[width=\linewidth]{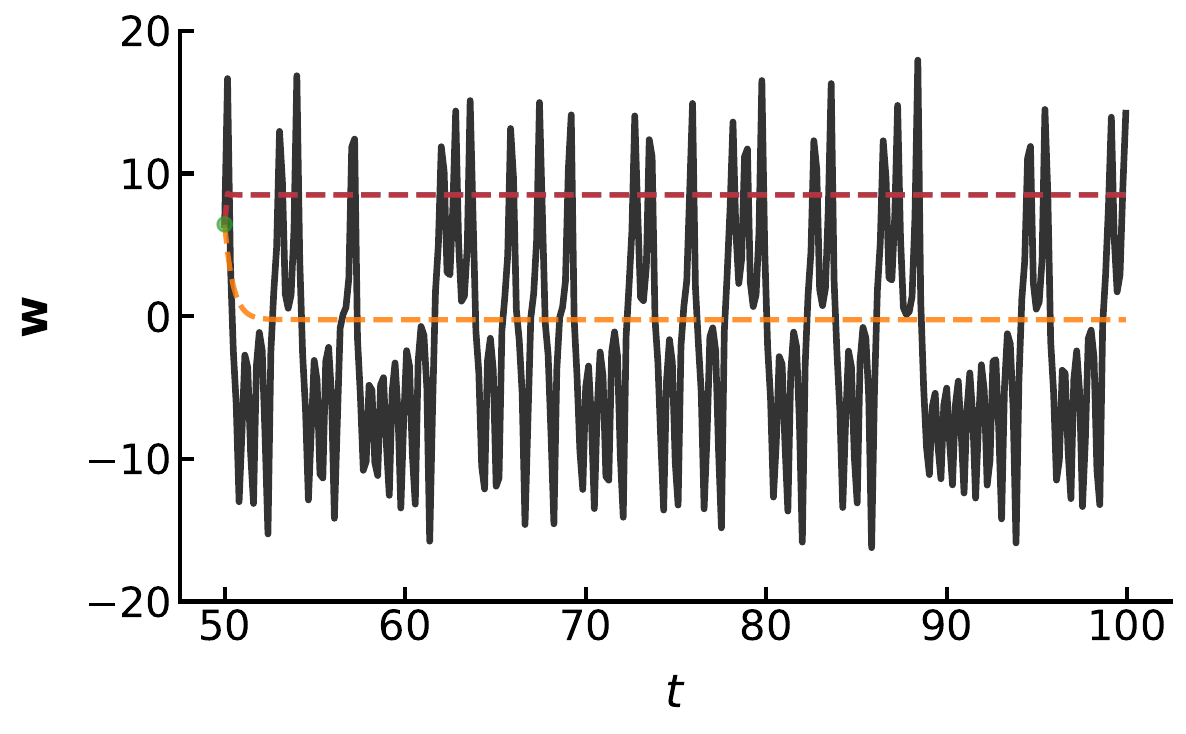}
        \caption{$n = 1$, $t\in (50, 100)$}
    \label{fig:lorenz63_n=1}
    \end{subfigure}\hfill
    \begin{subfigure}{0.49\textwidth}
        \includegraphics[width=\linewidth]{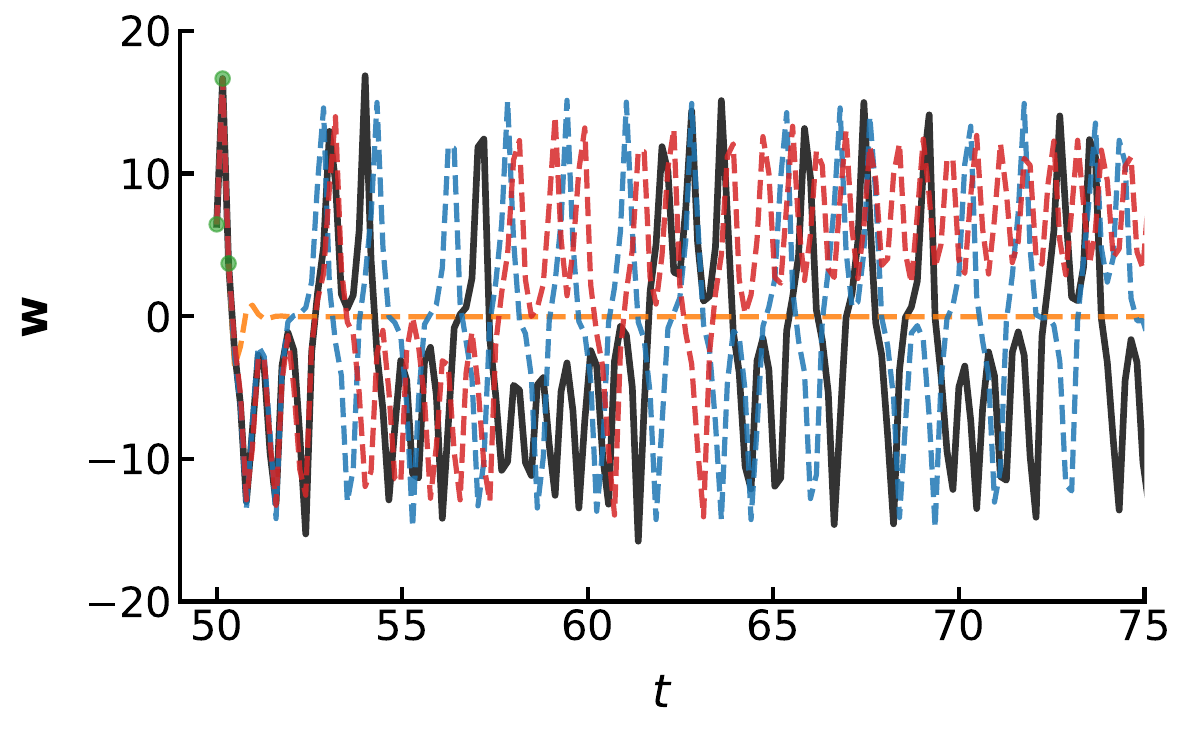}
        \caption{$n = 3$, $t\in (50, 75)$}
        \label{fig:lorenz63_n=3}
    \end{subfigure}
    \caption{A Lorenz '63 test trajectory with predictions from the TD-TF, TD-DMD and TD-MLP models, only observing the $x$-component of the trajectories.}
    \label{fig:lorenz63}
\end{figure}

For the TD-TF and the TD-MLP models, we set their hidden dimension to $100$, batch size to $100$, and learning rate to $10^{-2}$, and we train for $500$ epochs. We first assess the performance of the models by considering $n = 1$. Note that, in this case, TD-DMD reduces to the plain DMD, while the transformer reduces to a fully-connected residual network (ResNet) \citep{he2016deep}. As we can see in \Cref{fig:lorenz63_n=1}, none of the models can accurately reproduce the dynamics. 

We increase $n = 3$ in \Cref{fig:lorenz63_n=3} and notice that TD-DMD does not improve and still converges toward a trivial fixed point near zero. This is to be expected, since it is a fundamentally linear model and cannot represent the nonlinear mechanisms underlying chaotic dynamics. However, TD-TF now captures qualitatively the dynamics of the chaotic attractor. The TD-MLP captures some short-term oscillations, but it does not sustain the alternation between the two lobes of the chaotic attractor and instead settles into an approximately periodic orbit.

To give a more quantitative performance assessment of the TD-TF model on chaotic dynamics, we evaluate whether it reproduces simple statistical signatures of the Lorenz attractor. Specifically, for each test trajectory, we compute (i) the number of \emph{lobe switches}, defined as the number of sign changes in $x(t)$ between time intervals spent in the left and right lobes, and report the corresponding switching frequency obtained by dividing by the trajectory duration, and (ii) \emph{peak statistics} of the $x$-component, where a ``peak'' is a local maximum of $x(t)$ and $\Delta t_{\mathrm{peak}}$ denotes the time between successive peaks. We report the mean and standard deviation of these trajectory-wise quantities over $M=100$ test trajectories. The results are summarized in \Cref{tab:lorenz_stats}. TD-TF approximates both the switching rate and the oscillatory peak structure of the true system, although the larger standard deviations indicate substantial trajectory-to-trajectory variability in the long-horizon chaotic rollouts. This variability may be reduced by more systematic choices of initialization, learning-rate scheduling, and regularization. Nevertheless, the qualitative comparison is clear: TD-DMD is nearly non-switching, TD-MLP underestimates the switching rate on average, and TD-TF recovers the lobe-switching dynamics most accurately.

\begin{table}
\centering
\caption{Lobe-switching and peak statistics averaged over $M = 100$ test trajectories, for $n = 3$ and $h = 100$. To avoid spurious switches, we count a switch only after the trajectory satisfies $|x(t)|>0.1$. Means and standard deviations are computed over trajectories.}
\vspace{0.2cm}
\begin{tabular}{lcccc}
\hline
& \multicolumn{2}{c}{Lobe Switching} & \multicolumn{2}{c}{Peak Statistics} \\
\cline{2-3} \cline{4-5}
Model & Switches & Frequency & Peaks & Mean $\Delta t_{\mathrm{peak}}$ \\
\hline
True 
& $28.56 \pm 3.85$ 
& $0.5721 \pm 0.0770$
& $52.05 \pm 2.47$
& $0.9565 \pm 0.0451$
\\[0.1cm]
TD-DMD 
& $0.43 \pm 0.89$
& $0.0086 \pm 0.0177$
& $1.23 \pm 1.15$
& $0.7352 \pm 0.0700$
\\[0.1cm]
TD-TF
& $27.62 \pm 13.49$
& $0.5532 \pm 0.2701$
& $45.99 \pm 8.65$
& $1.1217 \pm 0.1721$
\\
TD-MLP
& $8.44 \pm 5.71$
& $0.1691 \pm 0.1144$ 
& $65.12 \pm 5.03$   
& $0.7764 \pm  0.0680$
\\
\hline
\end{tabular}
\label{tab:lorenz_stats}
\end{table}

\subsection{Case 4: Reaction--diffusion Partial Differential Equation}

As a final demonstration, we consider a spatially extended nonlinear reaction--diffusion system originally introduced in \citep{PhysRevLett.68.714}. 
This example bridges low-dimensional chaotic dynamics and high-dimensional PDE dynamics and tests the ability of TD-TF to learn effective reduced dynamics obtained from data-driven model reduction.

\paragraph{Setup.}
We consider the following one-dimensional reaction--diffusion system on the spatial domain, $x \in [0,1]$:
\begin{equation}\label{eq:reacdiff}
\begin{cases}
\partial_t u = D \partial_{xx} u + \dfrac{1}{\varepsilon}(v - u^2 - u^3), \\
\partial_t v = D \partial_{xx} v - u + \alpha,
\end{cases}
\end{equation}
with diffusion coefficient $D = 0.0322307$ and parameters $\varepsilon = \alpha = 0.01$.
This system arises as a prototypical activator--inhibitor model, where $u$ acts as a fast activator variable and $v$ as a slower recovery variable.
The cubic nonlinearity in $u$ produces bistability and excitable dynamics, while diffusion induces spatial coupling and pattern formation.
The chosen parameter values correspond to a regime exhibiting coherent spatio-temporal oscillations and low-dimensional dominant dynamics \citep{PhysRevLett.68.714}.
\begin{figure}[h]
    \centering
    \includegraphics[width = 0.7\textwidth]{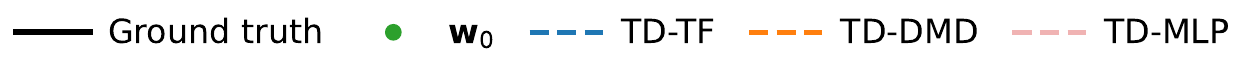}
    \includegraphics[width=0.65\linewidth]{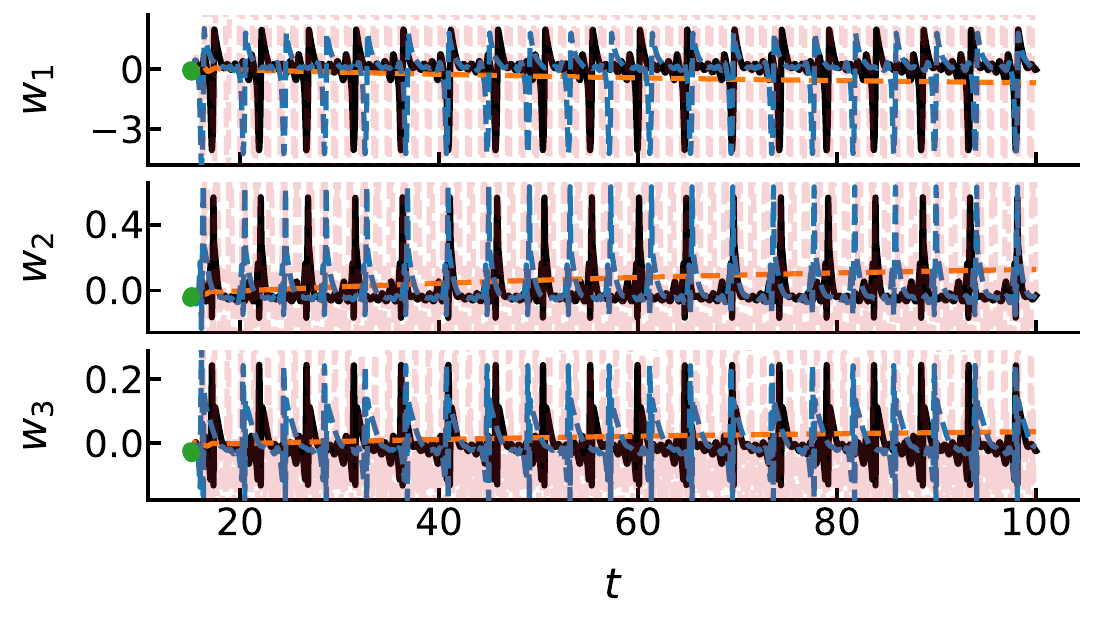}
    \caption{Trajectory of the reaction--diffusion case against the predictions of the TD-TF, TD-DMD and TD-MLP models, where $n=10$.}
    \label{fig:reacdiff_pod_time}
\end{figure}

We impose Dirichlet boundary conditions $u(0,t) = u(1,t) = -2, v(0,t) = v(1,t) = -4$, and solve the PDE using a spectral method \citep{burns2020dedalus}. In particular, we discretize in space using a Chebyshev basis with $N_x = 512$ modes and evolve the solution in time using a Runge--Kutta scheme with timestep $\Delta t = 10^{-4}$ up to final time $T = 100$. Data up to time $15$ is discarded to remove transient behavior.

From the resulting space--time field $u(x,t)$, we compute a proper orthogonal decomposition (POD) \citep{Holmes2012}.
We retain the first 3 POD modes ($d = 3$), which capture the dominant energetic structures of the dynamics. Projecting onto this basis yields a single trajectory ($N = 1$) of POD coefficients of length $K = 34000$, which is subsampled in time with $\tau = 5$. We construct $J = 2000$ burst samples using a delay of $n = 10$.

\paragraph{Results.}
We compare TD-DMD, TD-TF and TD-MLP on the task of long-term prediction of the POD coefficient dynamics. For both neural network models, we use a hidden dimension of $h = 100$, batch size of $100$, learning rate of $10^{-2}$, and weight decay of $10^{-5}$, and training is executed for $500$ epochs.

\Cref{fig:reacdiff_pod_time} shows the prediction of each POD coefficient for a representative trajectory.
TD-DMD exhibits rapid decay and loss of phase information, consistent with the limitations of linear models for nonlinear reduced dynamics, whereas TD-MLP produces high-frequency oscillations with incorrect amplitude and phase. In contrast, TD-TF accurately tracks the oscillatory structure of all three POD modes over long time horizons. 

\begin{figure}
    \centering

    \begin{subfigure}[t]{0.29\linewidth}
        \centering
        \includegraphics[width=\linewidth]{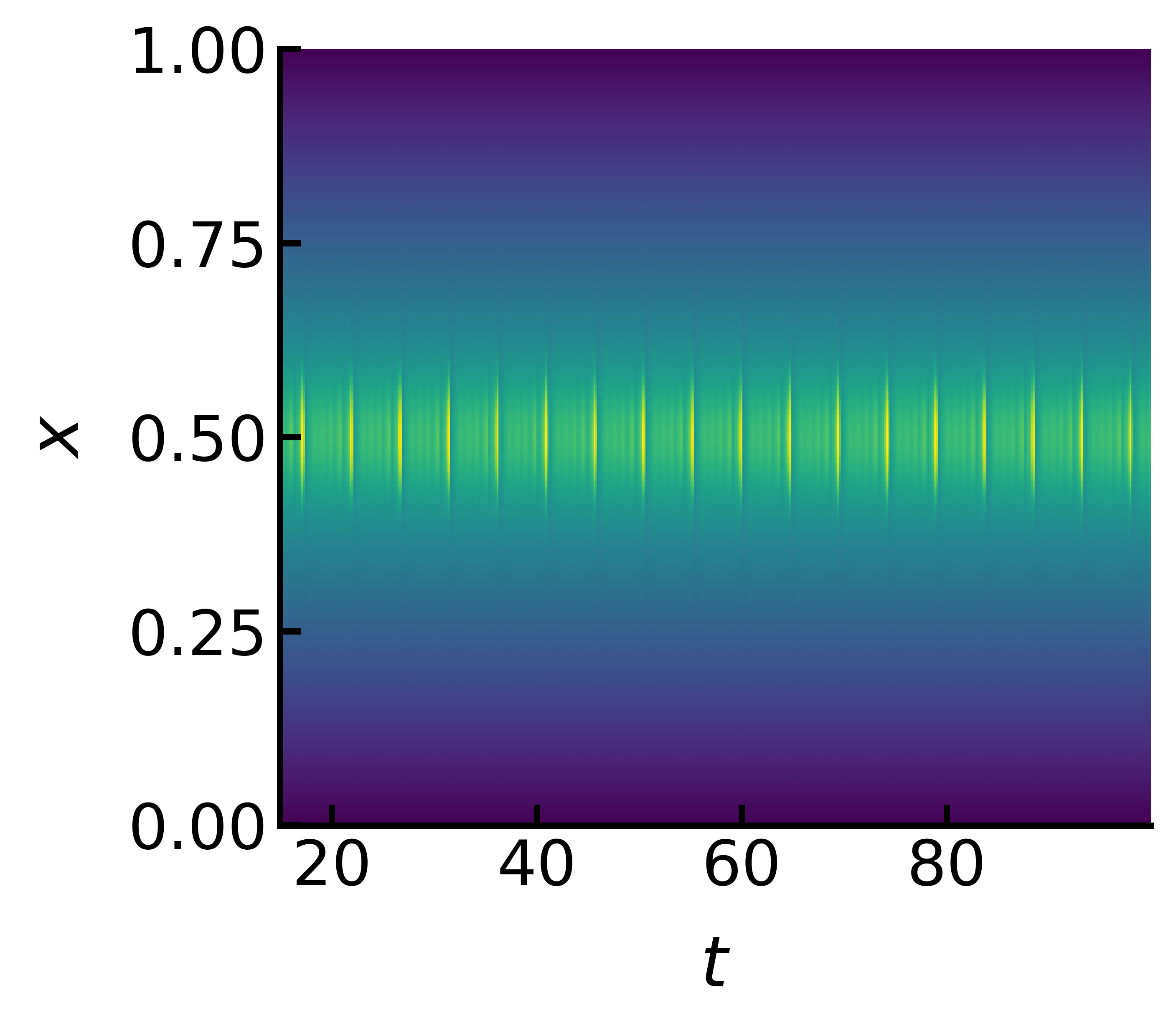}
        \captionsetup{margin={1.2em,0em}}
        \caption{Ground truth}
        \label{fig:reacdiff_recon_gt}
    \end{subfigure}
    \hfill
    \begin{subfigure}[t]{0.29\linewidth}
        \centering
        \includegraphics[width=\linewidth]{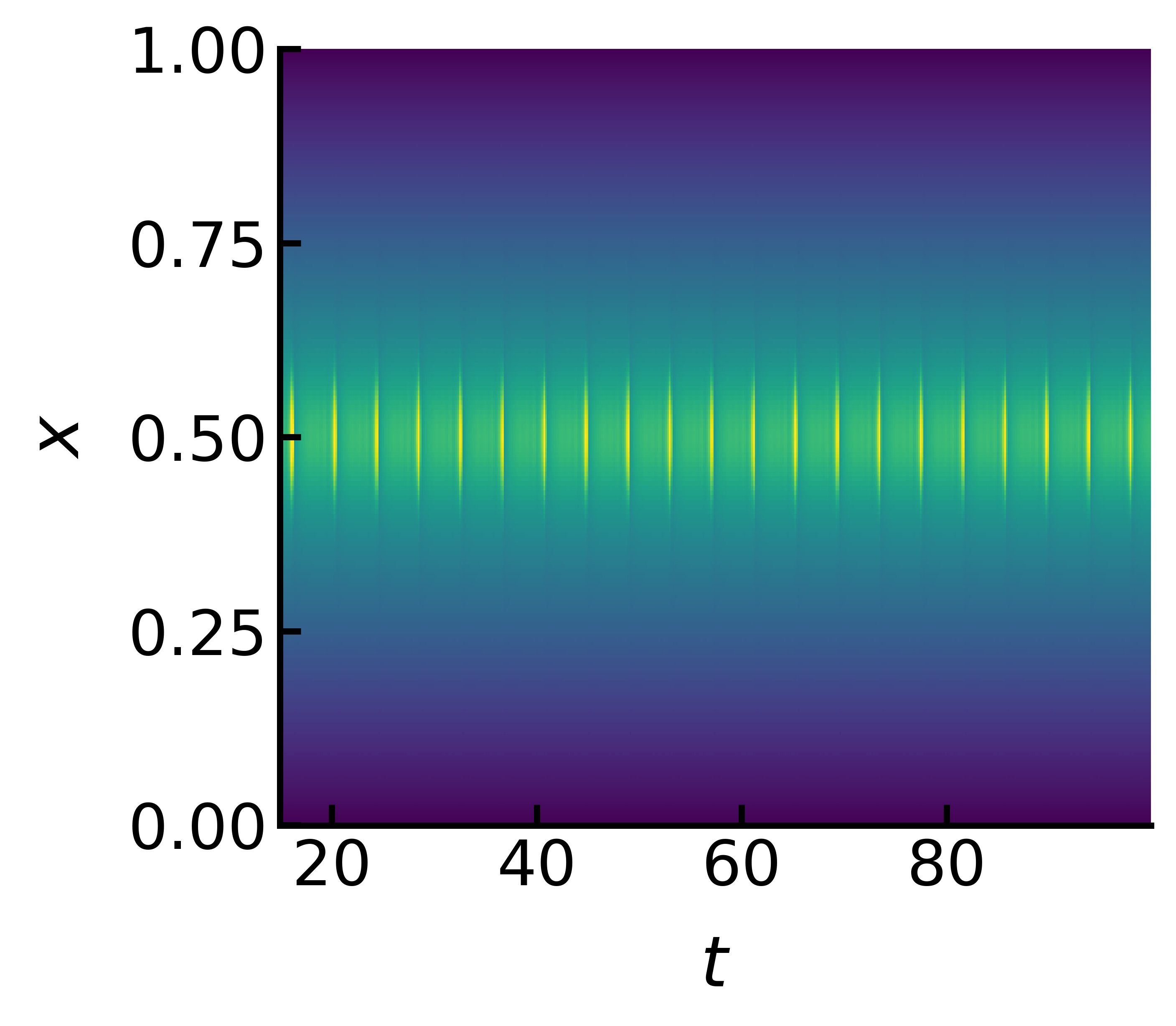}
        \captionsetup{margin={1.2em,0em}}
        \caption{TD-TF}
        \label{fig:reacdiff_recon_tdtf}
    \end{subfigure}
    \hfill
    \begin{subfigure}[t]{0.29\linewidth}
        \centering
        \includegraphics[width=\linewidth]{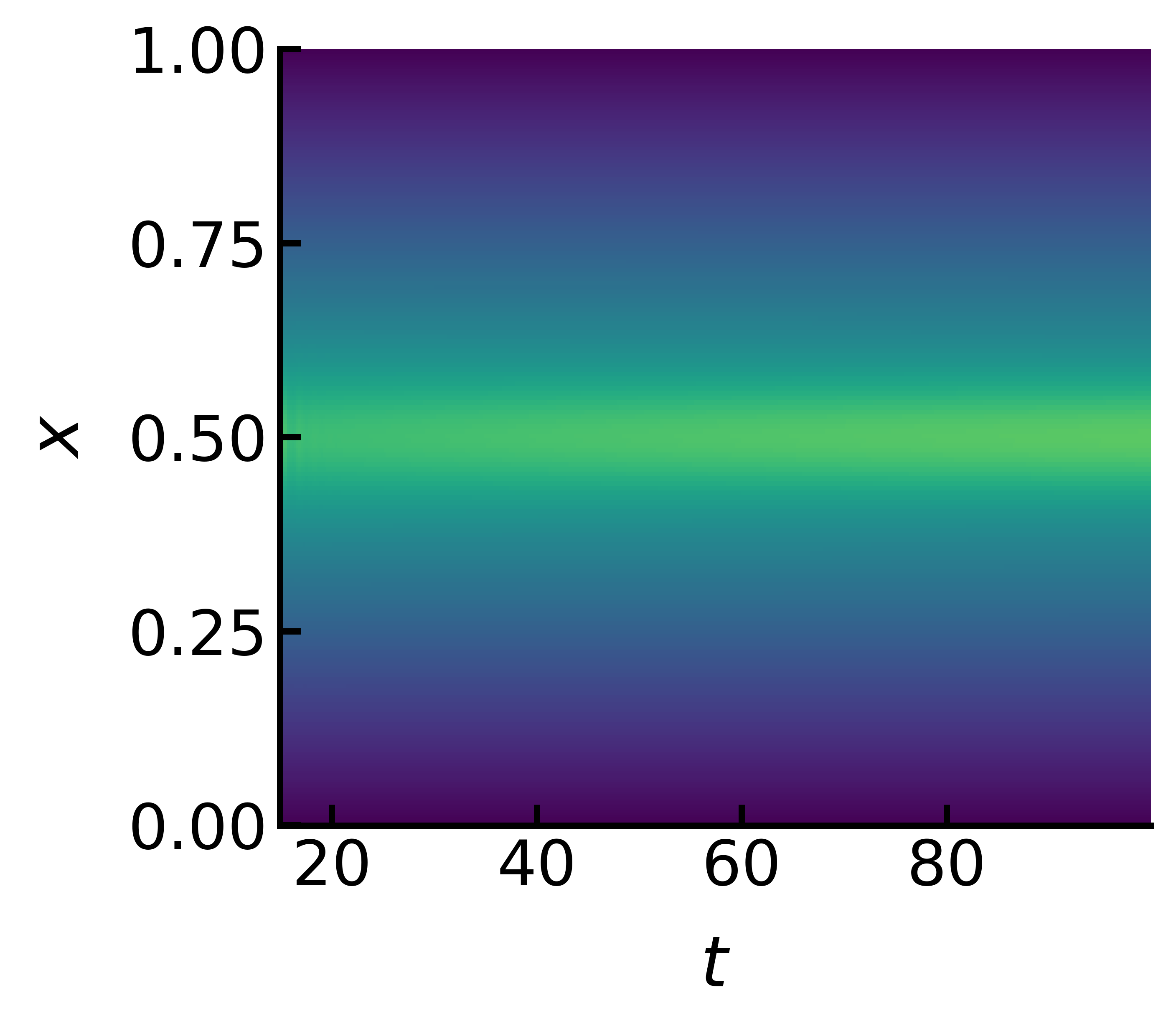}
        \captionsetup{margin={1.2em,0em}}
        \caption{TD-DMD}
        \label{fig:reacdiff_recon_tddmd}
    \end{subfigure}
    \hfill
    \begin{subfigure}[t]{0.074\linewidth}
        \centering
        \raisebox{0.64cm}{%
            \includegraphics[
                width=\linewidth,
                clip
            ]{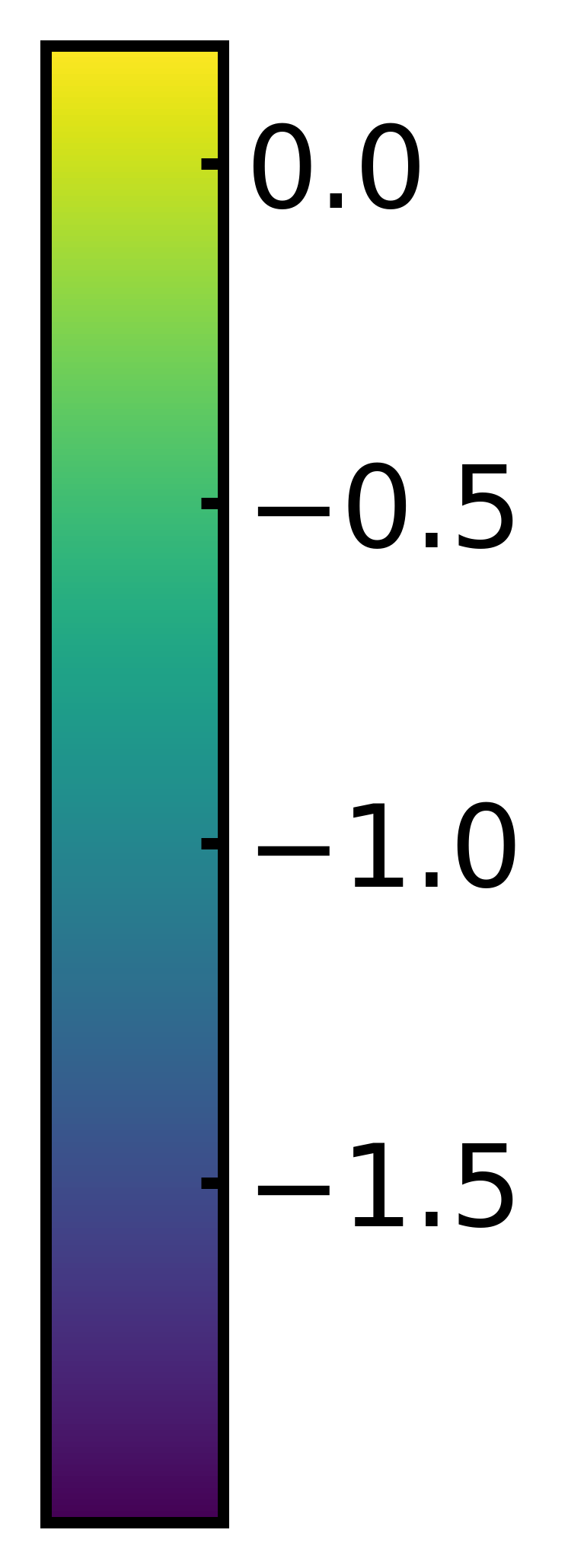}%
        }
    \end{subfigure}
    \caption{Reconstruction of the reaction--diffusion solution $u(x,t)$ from the first three POD modes.}
    \label{fig:reacdiff_recon}
\end{figure}

To assess how errors in the reduced dynamics propagate to the physical variables, we reconstruct the solution $u(x,t)$ from the predicted POD coefficients using the retained POD modes. \Cref{fig:reacdiff_recon} shows the ground truth reconstruction together with the reconstructions achieved by using the TD-TF and TD-DMD models for the same trajectory; TD-MLP is omitted here due to its unstable behavior.

The reconstruction achieved by using the TD-TF model reproduces the dominant spatio-temporal oscillatory patterns observed in the reference solution over the prediction horizon, maintaining comparable amplitudes and spatial organization, though with a clear phase misalignment. In contrast, the reconstruction by TD-DMD exhibits a progressive reduction in oscillation amplitude and a loss of spatial structure over time, consistent with the limitations of linear latent dynamics in representing nonlinear interactions among POD modes.

\begin{figure}[h]
    \centering
    \includegraphics[width=0.7\textwidth]{figs/reacDiff_legend.pdf}
    \vspace{0.5em}
    \begin{subfigure}{0.48\textwidth}
        \centering
        \includegraphics[width=\linewidth]{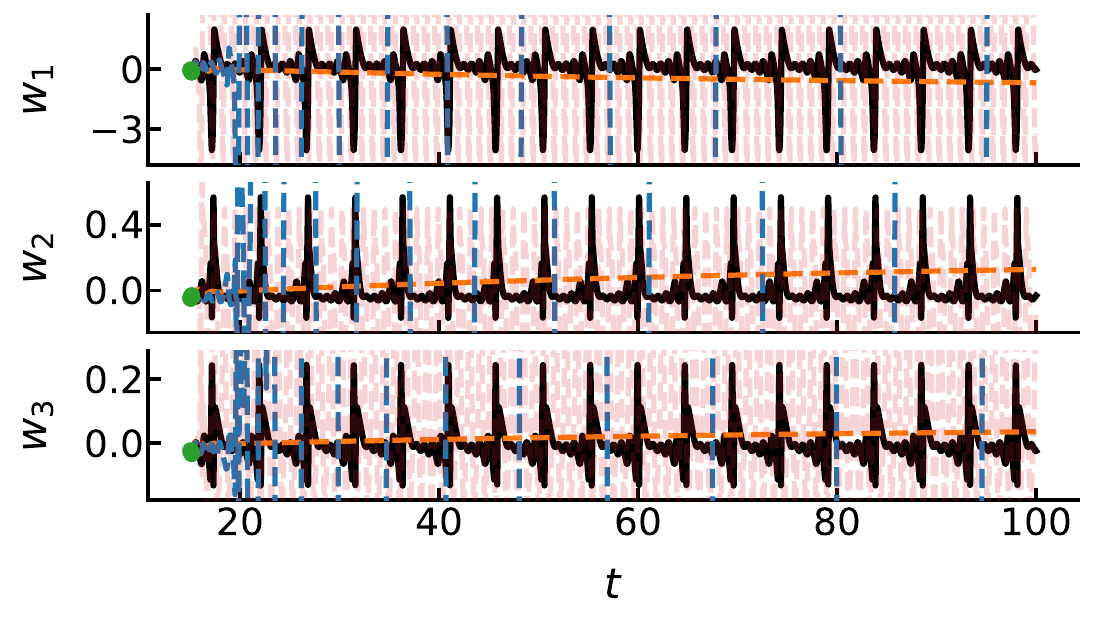}
    \end{subfigure} 
    \hfill
    \begin{subfigure}{0.48\textwidth}
        \centering
        \includegraphics[width=\linewidth]{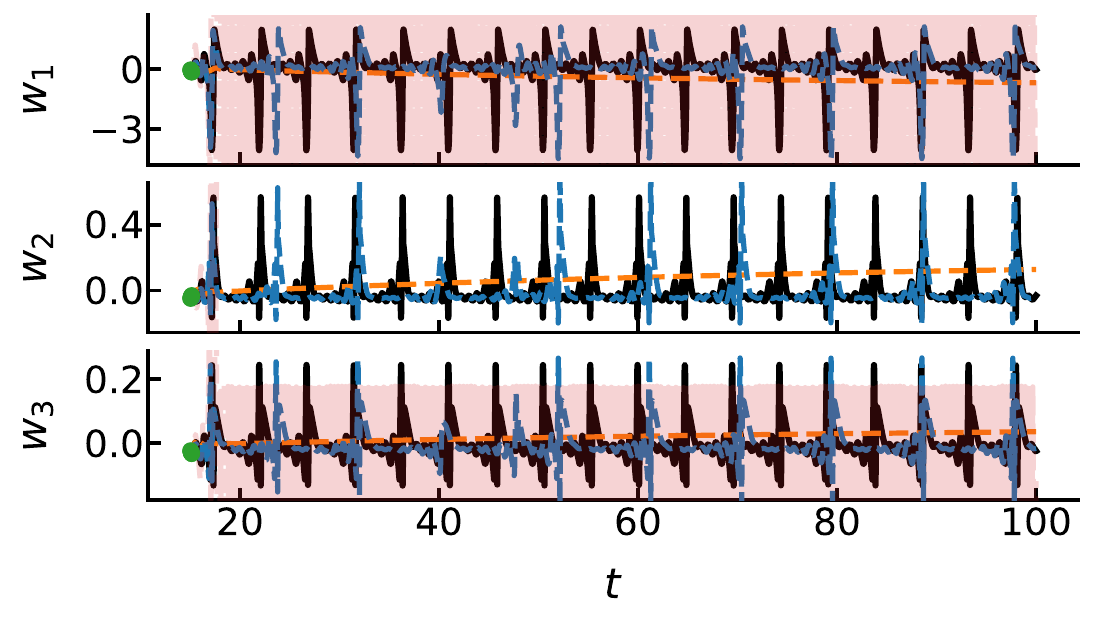}
    \end{subfigure}
    \caption{Reaction--diffusion trajectories and the corresponding predictions of the TD-TF, TD-DMD, and TD-MLP models for two different random seeds, with $n=10$ and $h=100$.}
    \label{fig:reacdiff_seed_sensitivity}
\end{figure}

Results for two additional seeds in \Cref{fig:reacdiff_seed_sensitivity} show that TD-TF is sensitive to random initialization in this setting, with visible variation in rollout quality across seeds, while TD-MLP exhibits similar failure modes across different seeds. Although seed-specific hyperparameter tuning may improve individual runs, improving robustness to initialization remains an important direction for future work.

\paragraph{Summary of the model comparison.}
The experiments show that TD-TF does not uniformly outperform TD-DMD. TD-DMD remains competitive, and can be preferable, when the dynamics are well approximated by a linear recurrence in delay coordinates, as in the sinusoidal case and parts of the NLR airfoil extrapolation study. TD-TF provides a clearer advantage when the delayed evolution map is nonlinear, as in the Lorenz system and the reaction--diffusion reduced dynamics. The TD-MLP comparisons indicate that these gains are not solely due to adding a nonlinear map on delay coordinates; the adaptive attention weighting seems to be important in the strongly nonlinear examples.

\section{Conclusions}\label{s:conclusions}

In this work, we have established a theoretical and practical connection between time-delayed linear models such as TD-DMD and the transformer architecture. By introducing TD-TF, we have demonstrated that such a simplified transformer can be interpreted as a nonlinear extension of TD-DMD, thereby bridging two previously separate modeling paradigms and enhancing interpretability. TD-TF retains the core idea of TD-DMD (each prediction arises from a finite window of past states) while leveraging the expressive power of attention to model nonlinear interactions among those states.  This attention mechanism is key to adaptively identifying and weighting the most relevant delayed states for each prediction.

Through numerical experiments on benchmark dynamical systems, we have shown that TD-TF does not simply replace TD-DMD; rather, it complements it. The method is most useful when the delayed evolution map is strongly nonlinear, while TD-DMD remains highly competitive, and sometimes preferable, in linear or weakly nonlinear regimes. 
These results suggest that time-delayed transformers might benefit from a design and simplification inspired by their linear counterparts. Our findings highlight that linking classical data-driven models to modern transformer architectures can enhance their interpretability. An appealing aspect of the TD-TF approach is its computational efficiency. The architecture uses only $O(n)$ attention operations per step, making training and deployment feasible even as sequence length grows large.

TD-TF represents a step toward trustworthy machine learning for dynamic systems. It combines the transparency of classical delay-embedded models with the power of modern transformers to handle nonlinearity. By respecting the fundamental delay structure, it ensures that only a finite history influences the future and that each past state's influence is explicit. We believe that this approach opens new avenues for data-driven modeling in engineering and science, enabling practitioners to build models that yield powerful yet interpretable predictions.

Future work will explore multi-dimensional and spatially extended systems (e.g. 2D fluid flows using TD-TF on latent variables), as well as incorporating multiple self-attention layers or heads to capture hierarchies of time-scales. Another avenue is enforcing physical constraints in TD-TF, for instance, via symmetry constraints, to further improve physical consistency.

%
%

\ack{The authors thank Giovanni Fantuzzi for valuable conversations. The authors gratefully acknowledge the scientific support and HPC resources provided by the German Aerospace Center (DLR).}

\funding{The work of A. Alcalde was funded by the European Union's Horizon Europe MSCA project ModConFlex (grant number 101073558).  The HPC system CARA is partially funded by ``Saxon State Ministry for Economic Affairs, Labor and Transport'' and ``Federal Ministry for Economic Affairs and Climate Action''.}

\roles{Albert Alcalde: Conceptualization, Methodology, Software, Validation, Visualization, Writing (original draft); Markus Widhalm: Conceptualization, Supervision, Data curation, Writing (review \& editing); Emre Y{\i}lmaz: Conceptualization, Supervision, Data curation, Software, Writing (review \& editing).}

\data{Benchmark datasets and code used in this study are available at \url{https://github.com/albertalcalde/TDTF}. The NLR7301 aerodynamic dataset is available from the authors upon reasonable request.}


\clearpage

\bibliographystyle{abbrv}  
\bibliography{refs}

\end{document}